\documentclass{article}
\usepackage[utf8]{inputenc}
\usepackage{fancyhdr}
\usepackage{arxiv}
\usepackage{graphicx}
\usepackage{tikz}
\usepackage{listings}
\usepackage{color}
\usepackage[pagebackref,breaklinks,colorlinks]{hyperref}
\usepackage{subcaption}
\usepackage{booktabs}
\usepackage{multirow}
\usepackage{bbm}
\usepackage{color, colortbl}
\usepackage[linesnumbered,ruled,vlined]{algorithm2e}
\usepackage{comment}
\usepackage{amsmath,amssymb} 
\usepackage{cite}
\usepackage{epsfig}
\usepackage{subcaption}
\usepackage{booktabs}
\usepackage{boldline}
\def\X{{\mathcal{X}}}
\def\Y{{\mathcal{Y}}}
\def\D{{\mathcal{D}}}

\def\dP{{\mathbb{P}}}
\def\cP{{\mathcal{P}}}

\begin{document}
\pagenumbering{arabic}
\title{Are Vision Transformers Robust to Spurious Correlations?} 

\author{Soumya Suvra Ghosal, Yifei Ming, Yixuan Li\\\\University of Wisconsin-Madison
\\ \texttt{\{sghosal, alvinming, sharonli\}@cs.wisc.edu}}
\date{}
\maketitle

\begin{abstract}
 Deep neural networks may be susceptible to learning spurious correlations that hold on average but not in atypical test samples. As with the recent emergence of vision transformer (ViT) models, it remains underexplored how spurious correlations are manifested in such architectures. In this paper, we systematically investigate the robustness of vision transformers to spurious correlations on three challenging benchmark datasets and compare their performance with popular CNNs. Our study reveals that when pre-trained on a sufficiently large dataset, ViT models are more robust to spurious correlations than CNNs. Key to their success is the ability to generalize better from the examples where spurious correlations do not hold. Further, we perform extensive ablations and experiments to understand the role of the self-attention mechanism in providing robustness under spuriously correlated environments. We hope that our work will inspire future research on further understanding the robustness of ViT models\footnote{Code is available at : \url{https://github.com/deeplearning-wisc/vit-spurious-robustness}}.  

\end{abstract}

\section{Introduction}
\label{sec:intro}

\begin{figure}[t]
\centerline{\includegraphics[scale = 0.5]{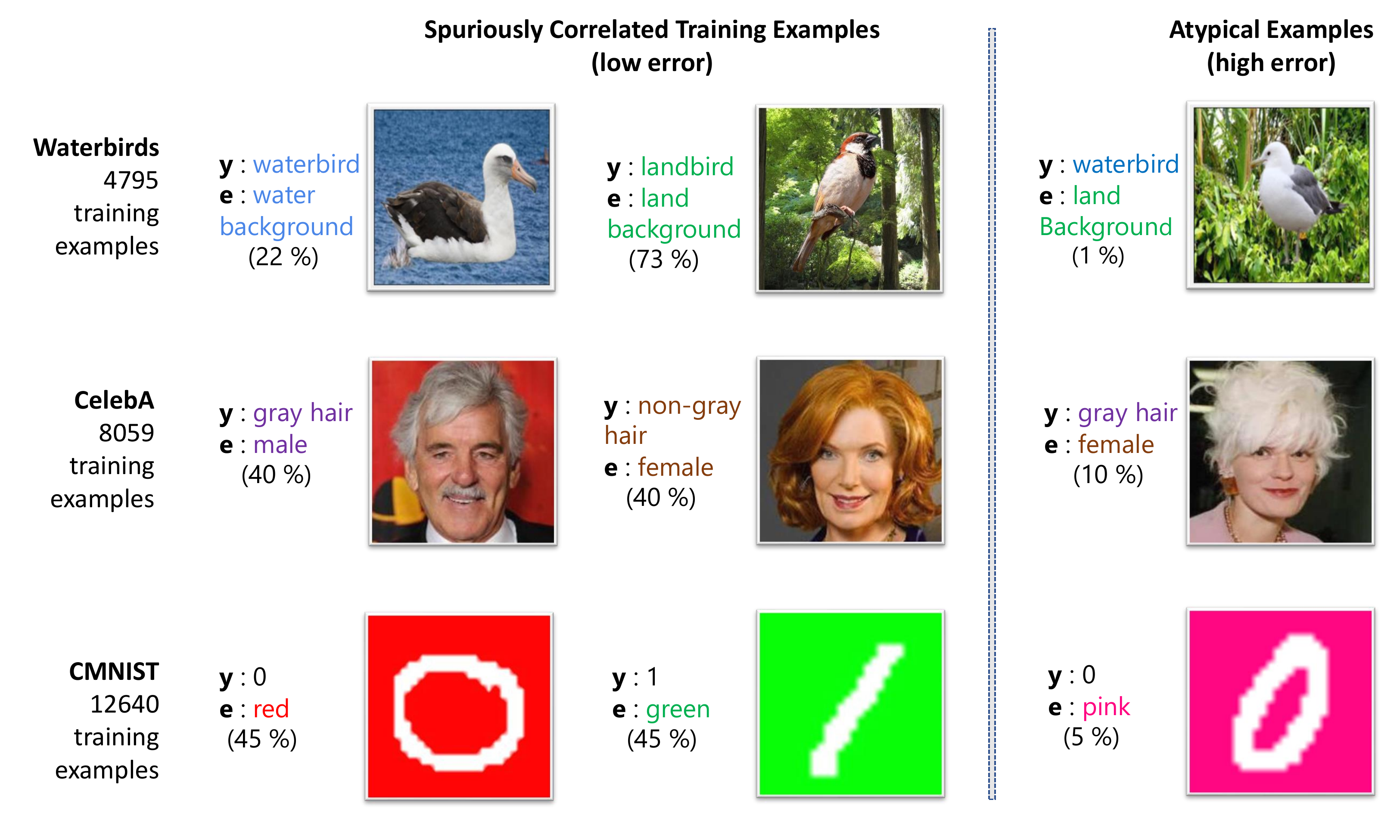}}
\caption{\textbf{Representative Examples.} We study three image datasets Waterbirds~\cite{sagawa2019distributionally}, CelebA~\cite{liu2015deep} and CMNIST. The label $y$ is spuriously correlated with environment $e$ in majority of training samples. The frequency of each group in training data is denoted by (\%). Figure is adapted from \cite{sagawa2019distributionally}.}

\label{fig:problem_setup}
\end{figure}

A key challenge in building robust image classification models is the existence of \emph{spurious correlations}: misleading heuristics imbibed within the training dataset that are correlated with majority examples but do not hold in general. 
Prior works have shown that convolutional neural networks (CNNs) can rely on spurious features to achieve high average test accuracy. Yet, such models lead to low accuracy on rare and untypical test samples lacking those heuristics~\cite{sagawa2019distributionally, geirhos2018imagenet,goel2020model, lifuempirical}. In Figure~\ref{fig:problem_setup}, we illustrate a model setup that exploits the spurious correlation between the \texttt{water background} and label \texttt{waterbird} for prediction. Consequently, a model that relies on spurious features performs poorly on test samples where the correlation no longer holds, such as \texttt{waterbird} on \texttt{land background}. 

While the robustness of CNNs has been widely studied, it remains underexplored how spurious correlation is manifested in the recent development of vision transformers (ViT)~\cite{dosovitskiy2020image}. As with the paradigm shift to attention-based architectures, it becomes increasingly critical to understand their behavior under ill-conditioned data. From a network architecture perspective, ViTs lack the inductive bias in CNNs, such as translational equivariance and spatial locality, and may be more prone to overfitting~\cite{dosovitskiy2020image}. For this reason, one may expect the fully-connected dependencies in ViT models may exacerbate capturing the spurious correlations in the training data. In this paper, we seek to answer the following question: \emph{Are Vision Transformers more robust to spurious correlations compared to CNNs}? 
Motivated by the question, we systematically investigate how and when ViT models exhibit robustness to spurious correlations on challenging benchmarks. Our findings reveal that for transformers, larger models and more pre-training data yield a significant improvement in robustness to spurious correlations. The key reason for success can be attributed to the ability to generalize better from those examples where spurious correlations do not hold, while fine-tuning. However, despite better generalization capability, ViT models suffer high errors on challenging benchmarks when these counterexamples are scarce in the training set. On the other hand, when pre-trained on a relatively smaller dataset such as ImageNet-1k, the performance of transformer-based models are much worse as compared to CNN counterparts. This indicates that in smaller pre-training data regimes, transformers have a higher propensity to overfit the spurious features and are less robust than CNNs of comparable size. 

Going beyond, we perform extensive ablations and experiments to understand the role of self-attention mechanism in providing robustness to ViT models. Our findings reveal that the self-attention mechanism in ViTs plays a crucial role in guiding the model to focus on spatial locations in an image which are essential for accurately predicting the target label. Interestingly, we also found that restricting the attention to be local can result in sharp degradation in model robustness to spurious associations. Thus, the global attention in ViT models is indeed important for providing additional robustness to spurious correlations. 

Our key contributions are summarized below: 
\begin{enumerate}
    \item To the best of our knowledge, we provide a first systematic study on the robustness of Vision Transformers when learned on datasets containing spurious correlations. Our work sheds light on the effectiveness of pre-training on ViT's robustness to spurious correlations. 
    \item We perform extensive experiments and ablations to understand the effect of model architectures, model capacity, pre-training dataset, data imbalance, fine-tuning, etc. 
    \item We provide insights on ViT's robustness by analyzing the attention matrix, which encapsulates important information about the interaction among image patches. We hope that our work will inspire future research on further understanding the robustness of ViT models.
  
\end{enumerate}

\section{Preliminaries}
\label{sec:prelims}

\subsection{Spurious Correlations}
\label{sec:spurious_def}


Spurious features refer to statistically informative features that work for majority of training examples but do not capture essential cues related
to the labels~\cite{sagawa2019distributionally, geirhos2018imagenet,goel2020model, lifuempirical}. We illustrate a few examples in Figure~\ref{fig:problem_setup}. In \texttt{waterbird} vs \texttt{landbird} classification problem, majority of the training images has the target label (\texttt{waterbird} or \texttt{landbird}) spuriously correlated with the background features (\texttt{water} or \texttt{land} background). 
Sagawa \textit{et al.}~\cite{sagawa2019distributionally} showed that deep neural networks can rely on these statistically informative yet spurious features to achieve high test accuracy on average, but fail significantly on groups where such correlations do not hold such as \texttt{waterbird} on \texttt{land background}.

Formally, we consider a training set, $\D^\text{train}$, consisting of $N$ training samples: $\{\*x_i, y_i\}^N_{i=1}$, where samples are drawn independently from a probability distribution: $\cP_{X,Y}$. Here, $X\in\X$ is a random variable defined in the pixel space, and $Y\in\Y = \{1,\ldots,K\}$ represents its label. We further assume that the data is sampled from a set of $E$ environments  $\mathcal{E} = \{e_1, e_2, \cdots, e_E\}$. The training data has spurious correlations, if the input $\*x_i$ is generated by a combination of invariant features $\*z^{inv}_i \in \mathbb{R}^{d_{inv}}$, which provides essential cues for accurate classification, and environmental features $\*z^{e}_i \in \mathbb{R}^{d_e}$ dependent on environment $e$: 
\vspace{-1mm}
\begin{align*}
    \*x_i = \rho(\*z^{inv}_i, \*z_i^e).
\end{align*}
Here $\rho$ represents a function transformation from the feature space $[\*z^{inv}_i, \*z_i^e]^T$ to the pixel space $\X$.
Considering the example of \texttt{waterbird} vs \texttt{landbird} classification, invariant features $\*z^{inv}_i$ would refer to signals which are essential for classifying $\*x_i$ as $y_i$, such as the feather color, presence of webbed feet, and fur texture of birds, to mention a few. Environmental features $\*z^{e}_i$, on the other hand, are cues not essential but correlated with target label $y_i$. For example, many waterbird
images are taken in water habitats, so water scenes can be considered as $\*z^{e}_i$. Under the data model, we form groups $g = (y,e) \in  \mathcal{Y}\times  \mathcal{E}$ that are jointly determined by the label $y$ and environment $e$. For this study, we consider the binary setting where $\mathcal{E} = \{1,-1\}$ and $\mathcal{Y} = \{1,-1\}$, resulting in four groups. The concrete meaning for each environment and label will be instantiated in corresponding tasks, which we describe in Section~\ref{sec:dataset}.


\subsection{Transformers}

Similar to the Transformer architecture in \cite{vaswani2017attention}, ViT model expects the input as a 1D sequence of token embeddings. An input image is first partitioned into non-overlapping fixed-size square patches of resolution $P \times P$, resulting in a sequence of flattened 2D patches. For example, given an image of size $384 \times 384$ and patch size $P=16$, the image is divided into patches of resolution $16 \times 16$, resulting in $576$ image patches. Next, these patches are mapped to constant size embeddings with a trainable linear projection. In the previous example, the output of the projection layer will be $576$ embedding vectors of fixed dimension. 

Following~\cite{devlin2018bert}, ViT prepends a learnable embedding (\texttt{class} token) to the sequence of embedded patches, and this \texttt{class} token is used as image representation at the output of the transformer. To imbibe relative positional information of patches, position embeddings are further added to the patch embeddings.

The core architecture of ViT mainly consists of multiple stacked encoder blocks, where each block primarily consists of: (1) multi-headed self-attention layers, which learn and aggregate information across various spatial locations of an image by processing interactions between different patch embeddings in a sequence; and (2) a feed-forward layer. See an expansive discussion in related work (Section~\ref{sec:related_works}).

\newcolumntype{?}{!{\vrule width 1pt}}

\setlength{\tabcolsep}{3pt}
\newcolumntype{M}[1]{>{\centering\arraybackslash}m{#1}}

\begin{table}[t]
 \renewcommand{\arraystretch}{1.5}
\centering
\resizebox{0.9\textwidth}{!}{%
 \begin{tabular}{c|cccc|ccc}
 
 \hlineB{2.5}
 \shortstack{\\\textbf{Pretraining}\\\textbf{Dataset}}  & \multicolumn{1}{c}{} & \multicolumn{1}{c}{} & \multicolumn{1}{c}{} & \multicolumn{1}{c}{} & \multicolumn{1}{c}{} &  \multicolumn{1}{c}{} \\[1ex]

\hline
\multirow{2}{*}{ImageNet-21k} & \textbf{Model} & ViT-B & ViT-S & ViT-Ti & BiT-M-R50x3 & BiT-M-R101x1 & BiT-M-R50x1 \\[2ex]
\cline{2-8}
& \textbf{\#Params} & 86.1M & 21.8M & 5.6M & 211M & 42.5M & 23.5M \\
\hline
\multirow{2}{*}{ImageNet-1k} & \textbf{Model} & DeiT-B & DeiT-S & DeiT-Ti & BiT-S-R50x3 & BiT-S-R101x1 & BiT-S-R50x1 \\[2ex]
\cline{2-8}
& \textbf{\#Params} & 86.1M & 21.8M & 5.6M & 211M & 42.5M & 23.5M \\
\hlineB{2}
\end{tabular}}
\vspace{0.2cm}
\caption{Different model architectures used in our experiments along with number of trainable parameters and pre-training dataset. Note that the DeiT architecture is identical to ViT variant of comparable size with the only difference lying in pre-training dataset and data augmentations used during pre-training.}
\vspace{-0.5cm}
\label{tab:architecture}
\end{table}
\vspace{-3mm}
\subsection{Model Zoo}
In this study, we aim to understand the robustness of ViT models when trained on a dataset containing spurious correlations and how they fare against popular CNNs. We contrast ViT with Big Transfer (BiT) models~\cite{Kolesnikov2020BigLearning} that are primarily based on the ResNet-v2 architecture. For both ViT and BiT models, we consider different variants that differ in model capacity and pre-training dataset, as summarized in Table~\ref{tab:architecture}. Specifically, we use model variants pre-trained on both ImageNet-1k~\cite{russakovsky2015imagenet} and on ImageNet-21k~\cite{deng2009imagenet} datasets. 

Table~\ref{tab:architecture} summarizes the size and pre-training dataset of different models used in our study. Note that the DeiT architecture is identical to ViT variant of comparable size with the only difference lying in the pre-training dataset and data augmentations.

\vspace{0.2cm}
\noindent \textbf{Notation:} To indicate input patch size in ViT models, we append ``/x'' to model names. We prepend -B, -S, -Ti to indicate \texttt{Base}, \texttt{Small} and \texttt{Tiny} version of the corresponding architecture. For instance: ViT-B/16 implies the \texttt{Base} variant with an input patch resolution of $16 \times 16$. In this paper, we use a $16 \times 16$ input patch size for computational simplicity.  

\section{Robustness to Spurious Correlation}
\label{sec:dataset}

In this section, we systematically measure the robustness performance of ViT models when trained on datasets containing spurious correlations, and compare how their robustness fares against popular CNNs. For evaluation benchmarks, we adopt the same setting as in~\cite{sagawa2019distributionally}. Specifically, we consider the following three classification datasets to study the robustness of ViT models in a spurious correlated environment: Waterbirds (Section~\ref{sec:waterbirds}), CelebA (Section~\ref{sec:celeba}), and ColorMNIST. Due to space constraints, results on ColorMNIST are in the Supplementary. 

\subsection{Waterbirds}
\label{sec:waterbirds}
Introduced in~\cite{sagawa2019distributionally}, this dataset contains spurious correlation between the background features and target label $y\in$ \{\texttt{waterbird}, \texttt{landbird}\}. The dataset is constructed by selecting bird photographs from the Caltech-UCSD Birds-200-2011 (CUB)~\cite{WahCUB_200_2011} dataset  and then superimposing on either of $e\in \mathcal{E}=  \{\texttt{water}, \texttt{land}\}$ background selected from the Places dataset~\cite{zhou2017places}. The spurious correlation is injected by pairing \texttt{waterbirds} on \texttt{water} background and \texttt{landbirds} on \texttt{land} background more frequently, as compared to other combinations. The dataset consists of $n = 4795$ training examples, with the smallest group size 56 (i.e, waterbird on land background).

\newcolumntype{?}{!{\vrule width 1pt}}
\renewcommand{\arraystretch}{1.25}
\begin{table}[t]
\centering
\resizebox{0.8\textwidth}{!}{%
 
 \begin{tabular}{l|c|c?c|c}
 
 \hlineB{2.5}
 \multirow{2}{0.75cm}{\centering Model} & \multicolumn{2}{c}{Train} & \multicolumn{2}{c}{Test} \\
 \cline{2-5}
 & Average Acc. & Worst-Group Acc. & Average Acc. & Worst-Group Acc. \\
\hline
 ViT-B/16 & 100 & 100 & \textbf{96.75} $\pm$ \textbf{0.05} & \textbf{89.30} $\pm$ \textbf{1.95} \\

 ViT-S/16 & 100 & 100 & 96.30 $\pm$ 0.51 & 85.45 $\pm$ 1.16 \\ 

 ViT-Ti/16 & 95.7 & 81.6 & 89.50 $\pm$ 0.05 & 71.65 $\pm$ 0.16 \\ 
 \hlineB{2}
 BiT-M-R50x3 & 100 & 100 & 94.90 $\pm$ 0.05 & 80.51 $\pm$ 1.02 \\ 

BiT-M-R101x1 & 100 & 100 & 94.05 $\pm$ 0.07 & 77.50 $\pm$ 0.50 \\ 

 BiT-M-R50x1 & 100 & 100 & 92.05 $\pm$ 0.05 & 75.10 $\pm$ 0.62 \\
 \hlineB{2.5}
 \end{tabular}}
 \vspace{0.2cm}
\caption{Average and worst-group accuracies over train and test set for different models when finetuned on Waterbirds~\cite{sagawa2019distributionally}. Both ViT-B/16 and ViT-S/16 attain better test worst-group accuracy as compared to BiT models. All models are pre-trained on ImageNet-21k. Results (mean and std) are estimated over 3 runs for each setting.}
\label{tab:waterbirds_avg_acc}
\vspace{-0.5cm}
\end{table}

\vspace{0.2cm}
\noindent \textbf{Results and insights on generalization performance} Table~\ref{tab:waterbirds_avg_acc} compares worst-group accuracies of different models when fine-tuned on Waterbirds~\cite{sagawa2019distributionally} using empirical risk minimization. Note that all the compared models are pre-trained on ImageNet-21k. This allows us to isolate the effect of model architectures, in particular, ViT vs. BiT models. The worst-group test accuracy reflects the model's generalization performance for groups where the correlation between the label $y$ and environment $e$ does not hold. A high worst-group accuracy is indicative of less reliance on the spurious correlation in training. Our results suggest that: (1) ViTs are relatively more robust to spurious associations between background feature and target label than convolution-based BiTs. Interestingly, ViT-B/16 attains a significantly higher worst-group test accuracy (89.3\%) than BiT-M-R50x3 despite having a considerably smaller capacity (86.1M vs. 211M). (2) Furthermore, these results reveal a correlation between generalization performance and model capacity. With an increase in model capacity, both ViTs and BiTs tend to generalize better, measured by both average accuracy and worst-group accuracy. The relatively poor performance of ViT-Ti/16 can be attributed to its failure to learn the intricacies within the dataset due to its compact capacity. 

\begin{figure}[t]
\centering
\begin{subfigure}{0.5\textwidth}
  \centering
  \includegraphics[width=.7\textwidth]{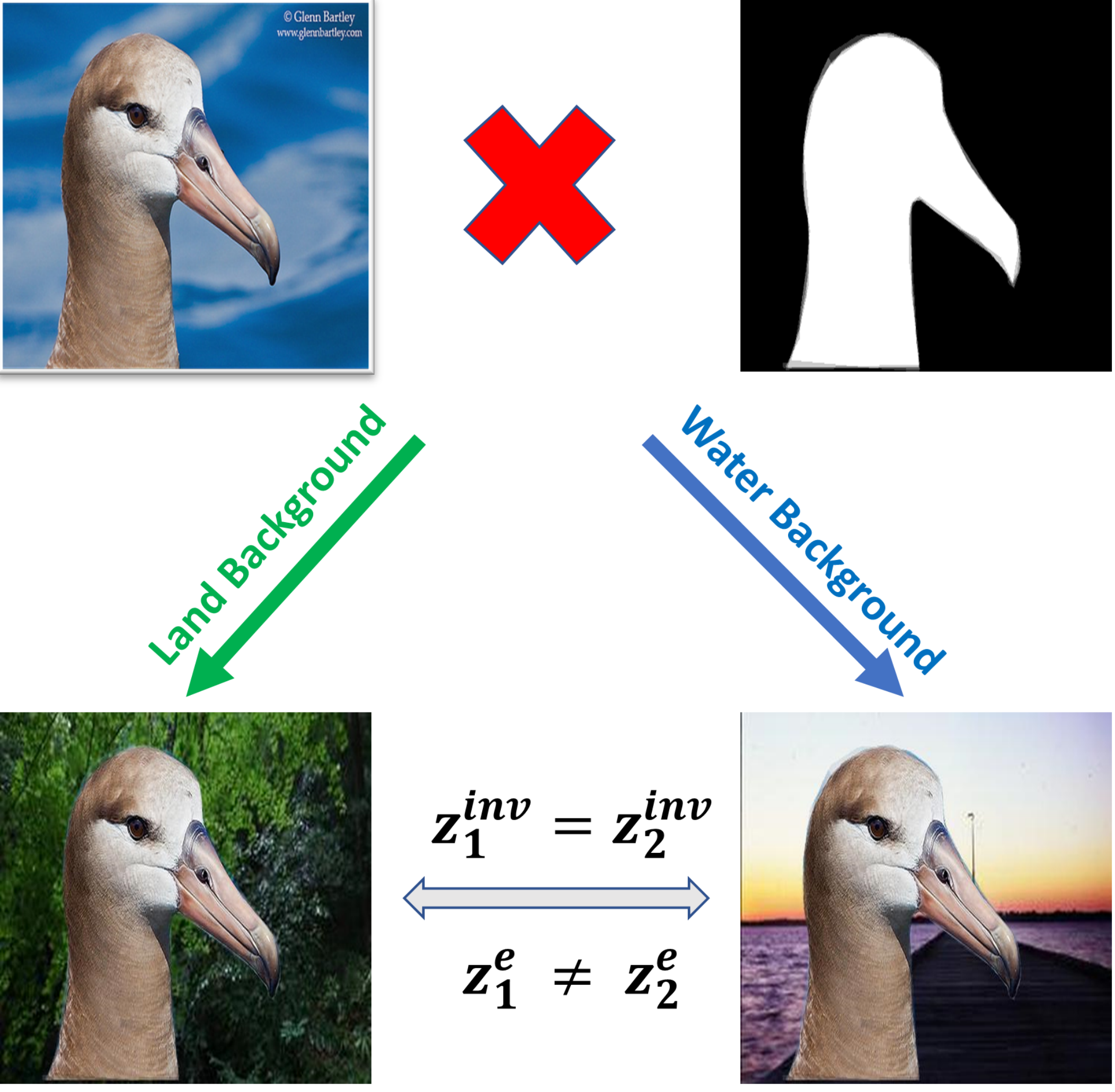}
  \label{fig:waterbirds_robustness_setup}
\end{subfigure}%
\begin{subfigure}{0.5\textwidth}
  \includegraphics[width=1\textwidth]{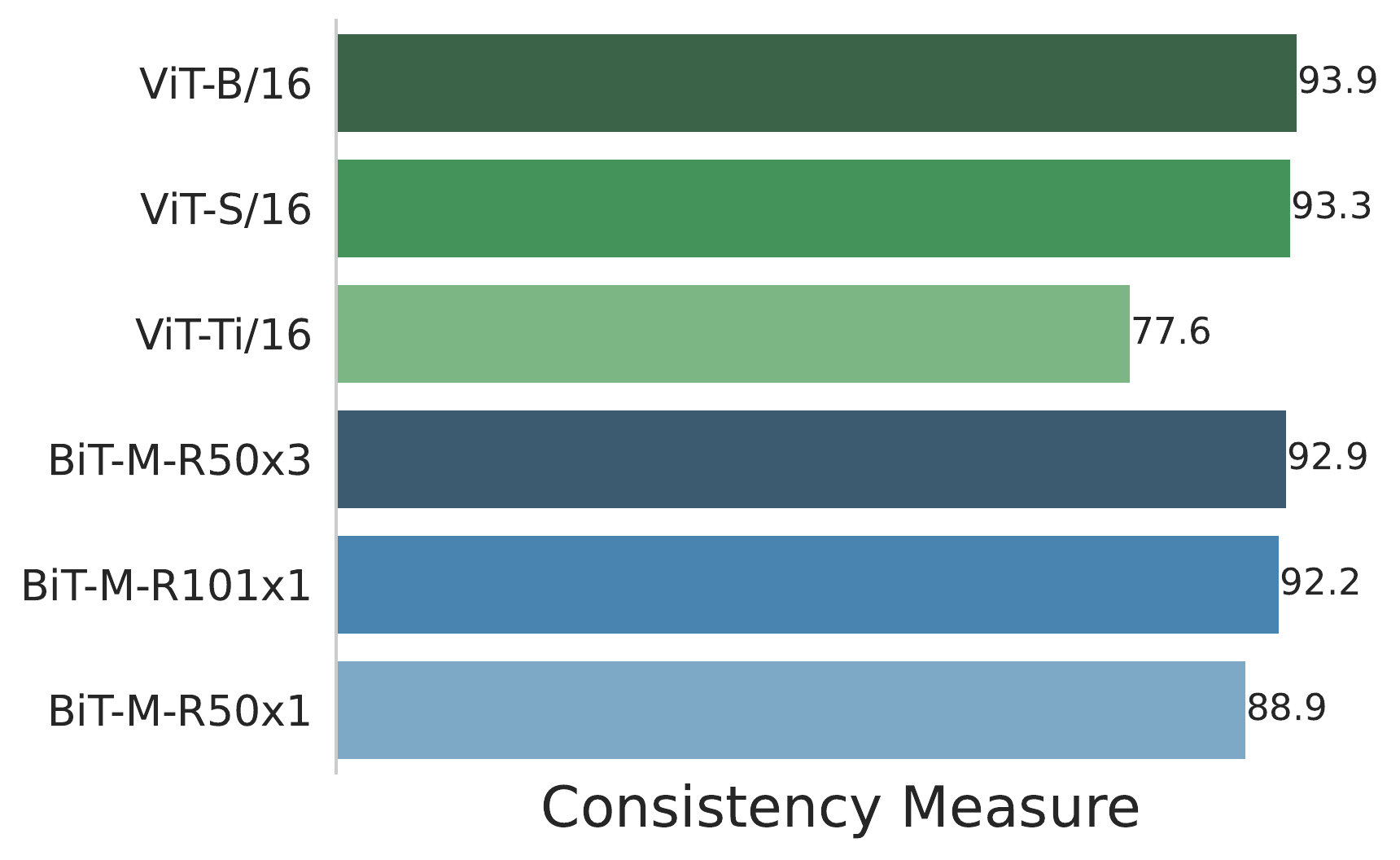}
  \label{fig:waterbirds_robustness_plot}
\end{subfigure}
\label{fig:waterbirds_robustness}
\caption{\textbf{Consistency Measure.} In Waterbirds dataset, y $\in$ \{\texttt{waterbird}, \texttt{landbird}\} is correlated with environment e $\in$ \{\texttt{water}, \texttt{land}\}. \textbf{Left}: Visual illustration of the experimental setup for measuring model consistency. Ideally, changing the spurious features ($\*z^e$) should have no impact on model prediction. \textbf{Right}: Evaluation results quantifying consistency for models of different architectures and varying capacity.} 
\label{fig:waterbirds_robustness}
\end{figure}

\noindent \textbf{Results and insights on robustness performance} We now delve deeper into the robustness of ViT models. In particular, we investigate the robustness in model prediction under varying background features. Our key idea is to compare the predictions of  image pairs ($\*x_i$, $\bar{\*x}_i$) with the same foreground object yet different background features (i.e., \texttt{water} vs. \texttt{land} background). We define \emph{Consistency Measure} of a model as the average number of consistent predictions on the evaluation dataset given the predictions are correct, i.e., $\frac{1}{N}\sum_{i = 1}^N \mathbb{I}{\{\hat{f}(\*x_i)= \hat{f}(\bar{\*x}_i) \mid \hat{f}(\*x_i) = y_i\}}$, where $y_i$ denotes the target label. To generate the image pairs ($\*x_i$, $\bar{\*x}_i$), we first take a foreground bird photograph using the pixel-level segmentation masks from the CUB dataset~\cite{WahCUB_200_2011}. We then place it on the top of  \texttt{water} and \texttt{land} background images  from the Places dataset~\cite{zhou2017places}. We generate multiple such pairs to form the evaluation dataset $\{(\*x_i, \bar{\*x}_i)\}_{i=1}^N$ and use this dataset to quantify the robustness performance. For this study, the evaluation dataset consists of $N=11788$ paired samples.

Figure~\ref{fig:waterbirds_robustness} provides a visual illustration of the experimental setup (\textbf{left}), along with the evaluation results (\textbf{right}). Our operating hypothesis is that a robust model should  predict same class label $\hat{f}(\*x_i)$ and ${\hat{f}}(\bar{\*x}_i)$  for a given pair $(\*x_i, \bar{\*x}_i)$, as they share exactly the same foreground object (i.e., invariant feature). Our results in Figure~\ref{fig:waterbirds_robustness} show that ViT models achieve overall higher consistency measures than BiT counterparts. For example, the best model ViT-B/16 obtains consistent predictions for 93.9\% of image pairs. Overall, using ViT pre-trained models yields strong generalization and robustness performance on Waterbirds. 

\subsection{CelebA}
\label{sec:celeba}

Beyond background spurious features, we further validate our findings on a different type of spurious feature based on gender attributes. Here, we investigate the behavior of machine learning models when learned on training samples with spurious associations between target label and demographic information such as gender.  Following~\cite{ming2022spurious}, we use CelebA dataset, consisting of celebrity images with each image annotated using 40 binary attributes. We have the label space $\mathcal{Y} = \{\texttt{gray hair}, \texttt{nongray hair}\}$ and gender as the spurious feature, $\mathcal{E} = \{\texttt{male}, \texttt{female}\}$. The training data consists of $4010$ images with label \texttt{grey hair}, out of which $3208$ are \texttt{male}, resulting in spurious association between gender attribute \texttt{male} and label \texttt{grey hair}. Formally, $\dP(e = \texttt{grey hair} | y = \texttt{male}) \approx \dP(e = \texttt{non-grey hair}  | y = \texttt{female} ) \approx 0.8$.

\newcolumntype{?}{!{\vrule width 1pt}}
\renewcommand{\arraystretch}{1.25}
\begin{table}[t]
\centering
\resizebox{0.80\textwidth}{!}{
 \begin{tabular}{l|c|c?c|c}
 \hlineB{2.5}
 \multirow{2}{0.75cm}{\centering Model} & \multicolumn{2}{c}{Train} & \multicolumn{2}{c}{Test} \\
 \cline{2-5}
 & Average Acc. & Worst-Group Acc. & Average Acc. & Worst-Group Acc. \\
  \hline
 ViT-B/16 & 100 & 100 & \textbf{97.40} $\pm$ \textbf{0.62} & \textbf{94.10} $\pm$ \textbf{0.51} \\

 ViT-S/16 & 100 & 100 & 96.26 $\pm$ 0.66 & 91.50 $\pm$ 1.56 \\ 

 ViT-Ti/16 & 97.9 & 93.3 & 96.71 $\pm$ 0.18 & 88.60 $\pm$ 3.92 \\ 
 \hlineB{2}
 BiT-M-R50x3 & 100 & 100 & 97.31 $\pm$ 0.05 & 89.80 $\pm$ 0.42 \\ 

 BiT-M-R101x1 & 100 & 100 & 97.20 $\pm$ 0.08 & 89.33 $\pm$ 0.78 \\ 

 BiT-M-R50x1 & 100 & 100 & 96.82 $\pm$ 1.2 & 87.72 $\pm$ 1.56 \\ 
 \hlineB{2.5}
 \end{tabular}}
 \vspace{0.2cm}
\caption{Average and worst-group accuracies over train and test set for different models when finetuned on CelebA~\cite{liu2015deep}. Both ViT-B/16 and ViT-S/16 attain better worst-group accuracy as compared to BiT models. All models are pre-trained on ImageNet-21k. Results (mean and std) are estimated over 3 runs for each setting.}

\label{tab:celeba_avg_acc}
\end{table}

\vspace{0.3cm}
\noindent \textbf{Results} We see from Table~\ref{tab:celeba_avg_acc} that ViT models achieve higher test accuracy (both average and worst-group) as opposed to BiTs. In particular, ViT-B/16 achieves $+\textbf{4.3}\%$ higher worst-group test accuracy than BiT-M-R50x3, despite having a considerably smaller capacity (86.1M vs. 211M). These findings along with our observations in Section~\ref{sec:waterbirds} demonstrate that ViTs are not only more robust when there are strong associations between the label and background features, but also avoid learning spurious correlations between demographic features and target label.

\section{Discussion: A Closer Look at ViT Under Spurious Correlation}
\label{sec:discussion}

\newcolumntype{?}{!{\vrule width 1pt}}
\renewcommand{\arraystretch}{1.25}
\setlength{\tabcolsep}{7pt}
\begin{table}[t]
\centering
\resizebox{0.8\textwidth}{!}{
 \begin{tabular}{cl|c?c|c}
 
 \hlineB{2.5}
 & \multirow{2}{0.75cm}{\centering Model} & \multicolumn{2}{c|}{Test Accuracy} & \multirow{2}{1.5cm}{\centering\shortstack{Consistency\\Measure$\pmb{\uparrow}$}}\\
 \cline{3-4}
&  & Average Acc. & Worst-Group Acc. & \\
  \hline
\multirow{3}{2cm}{ImageNet-21k} & ViT-B/16 & 96.8 & 89.3 & 93.9\\

& ViT-S/16 & 96.3 & 85.5 & 93.3 \\ 

& ViT-Ti/16 & 89.5 & 71.7 & 77.6\\ 
 \hlineB{2}
\multirow{3}{2cm}{ImageNet-1k} & DeiT-B/16 & 85.9 & 44.6  & 71.9\\ 

& DeiT-S/16 & 84.5 & 46.7 & 74.3\\ 

& DeiT-Ti/16 & 83.4 & 41.8 & 71.1\\ 
 \hlineB{2}
  \multirow{3}{2cm}{ImageNet-21k} & BiT-M-R50x3 & 94.9 & 80.5 & 92.9 \\
& BiT-M-R101x1 & 94.1 & 77.5 & 92.2\\ 
& BiT-M-R50x1 & 92.1 & 75.1 &  88.9\\ 
 \hlineB{2}
 \multirow{3}{2cm}{ImageNet-1k}& BiT-S-R50x3 & 87.0 & 60.3 &  77.8 \\ 
& BiT-S-R101x1 & 87.3 & 64.9 & 80.8 \\ 
& BiT-S-R50x1 & 86.3 & 63.5 &  78.7 \\ 
 \hlineB{2.5}\\
 \end{tabular}}
\caption{Investigating the effect of large scale pre-training on model robustness to spurious correlations. All models are fine-tuned on Waterbirds~\cite{sagawa2019distributionally}. Using ImageNet-21k for pre-training attains better performance.}
\label{tab:pretraining_vit}

\end{table}

In this section, we perform extensive ablations and experiments to understand the role of ViT models under spurious correlations. For consistency, we present the analyses below based on the Waterbirds dataset. 

\subsection{How does the size of the pre-training dataset affect robustness to spurious correlations?}

In this section, we aim to understand the role of large-scale pre-training on the model's robustness to spurious correlations. Specifically, we compare pre-trained models of different capacities, architectures, and sizes of pre-training data. To understand the importance of the pre-training dataset, we compare models pre-trained on ImageNet-1k ($1.3$ million images) and ImageNet-21k ($12.8$ million images). We report results for transformer-based models and BiT models in Table~\ref{tab:pretraining_vit}. For detailed ablation results on other benchmark datasets, please refer to the Appendix. Based on these results, we highlight the following observations:
\begin{enumerate}
    \item First, large-scale pre-training improves the performance of the models on challenging benchmarks. For transformers, larger models (\texttt{base} and \texttt{small}) and more pre-training data (ImageNet-21k) yields a significant improvement in all reported metrics. Hence, larger pre-training data and increasing model size play a crucial role in improving model robustness to spurious correlations. We also see a similar trend in the case of BiT models.
    
    \item Second, when pre-trained on a relatively smaller dataset such as ImageNet-1k, the performance of transformer-based DeiT models are much worse as compared to BiT-S models. Interestingly, although increasing size of DeiT models leads to improved average test accuracy but suffers high error on worst-group samples. This indicates that in smaller pre-training data regimes, transformers have a higher propensity of memorizing training samples and are less robust compared to CNNs of comparable size. From a network architecture perspective, this may be due to fully-connected layers in transformer models which capture spurious correlations occurring in the target task in case of limited pre-training data. Our findings corroborate reportings in~\cite{dosovitskiy2020image, bhojanapalli2021understanding} that inductive bias in convolutional neural networks plays a crucial role without strong pre-training. 
\end{enumerate}

\subsection{Understanding role of self-attention mechanism for improved robustness in ViT models}
\label{sec:attention}

Given the results above, a natural question arises: what makes ViT particularly robust in the presence of spurious correlations? In this section, we aim to understand the role of ViT by looking into the self-attention mechanism. The attention matrix in ViT models encapsulates crucial information about the interaction between different image patches. 

\begin{figure}[t]
\centerline{\includegraphics[scale = 0.5]{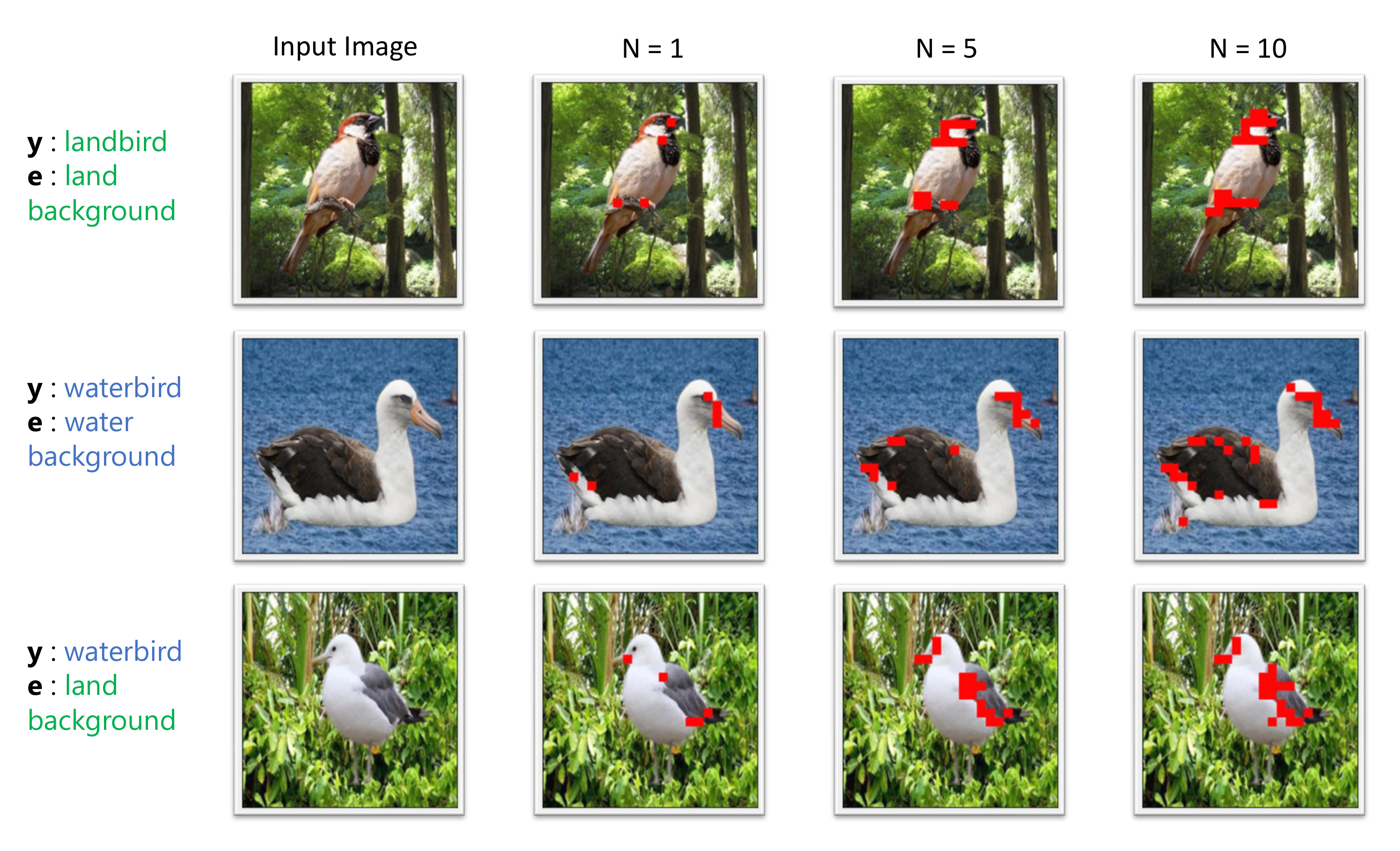}}
\caption{\small Visualization of the top $N$ patches receiving the highest attention (marked in \textcolor{red}{red}). Investigating the attention matrix, we find that all image patches---irrespective of spatial location---provides maximum attention to the patches representing essential cues for accurately identifying the foreground object such as claw, beak and fur color. See text for details. See Supplementary for visualizations on other datasets and models.}
\vspace{-0.5cm}
\label{fig:latent_pattern}
\end{figure}

\vspace{0.3cm}
\noindent \textbf{Latent pattern in attention matrix} To gain insights, we start by analyzing the attention matrix, where each element in the matrix $a_{i,j}$ represents attention values with which an image patch $i$ focuses on another patch $j$. For example: consider an input image of size $384 \times 384$ and patch resolution of $16 \times 16$, then we have a $576 \times 576$ attention matrix (excluding the \texttt{class token}). To compute final attention matrix, we use Attention Rollout~\cite{abnar-zuidema-2020-quantifying} which recursively multiplies attention weight matrices in all layers below. Our analysis here is based on the ViT-B/16 model fine-tuned on Waterbirds.

Intriguingly, we observe that each image patch, irrespective of its spatial location, provides maximum attention to the patches representing essential cues for accurately identifying the foreground object. 
 
Figure~\ref{fig:latent_pattern} exhibits this interesting pattern, where we mark (in \textcolor{red}{red}) the top $N=\{1,5,10\}$ patches being attended by every image patch.  To do so, for every image patch $i$, where $i \in \{1,\cdots,576\}$, we find the top $N$ patches receiving the highest attention values and mark (in \textcolor{red}{red}) on the original input image. This would give us $576 \times N$ patches, which we overlay on the original image. Note that different patches may share the same top patches, hence we observe the sparse pattern. In Figure~\ref{fig:latent_pattern}, we can see that the patches receiving the highest attention represent important signals such as the shape of the beak, claw, and fur color---all of which are essential for the classification task \texttt{waterbird} vs \texttt{landbird}.

It is particularly interesting to note the last row in Figure~\ref{fig:latent_pattern}, which is an example from the minority group (\texttt{waterbird} on \texttt{land} background). This is a challenging case where the spurious correlations between $y$ and $e$ do not hold. A non-robust model would utilize the background environmental features for predictions. In contrast, we notice that each patch in the image correctly attends to the foreground patches.

 \begin{figure}[t]
\centering
\begin{subfigure}{0.5\textwidth}
  \centering
  \includegraphics[width=0.9\textwidth]{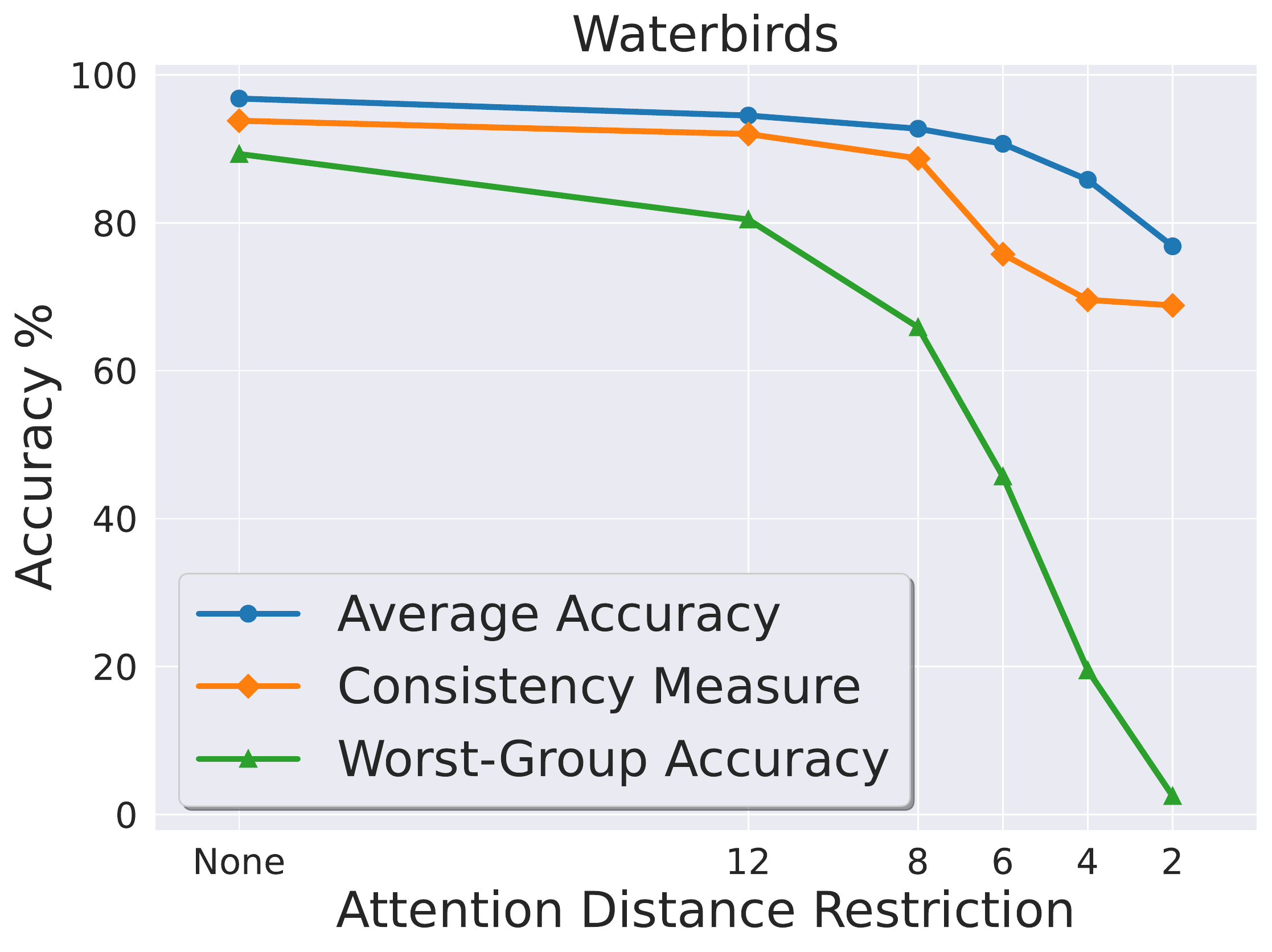}
  \label{fig:masked_attention_waterbirds}
\end{subfigure}%
\begin{subfigure}{0.5\textwidth}
 
  \includegraphics[width=0.9\textwidth]{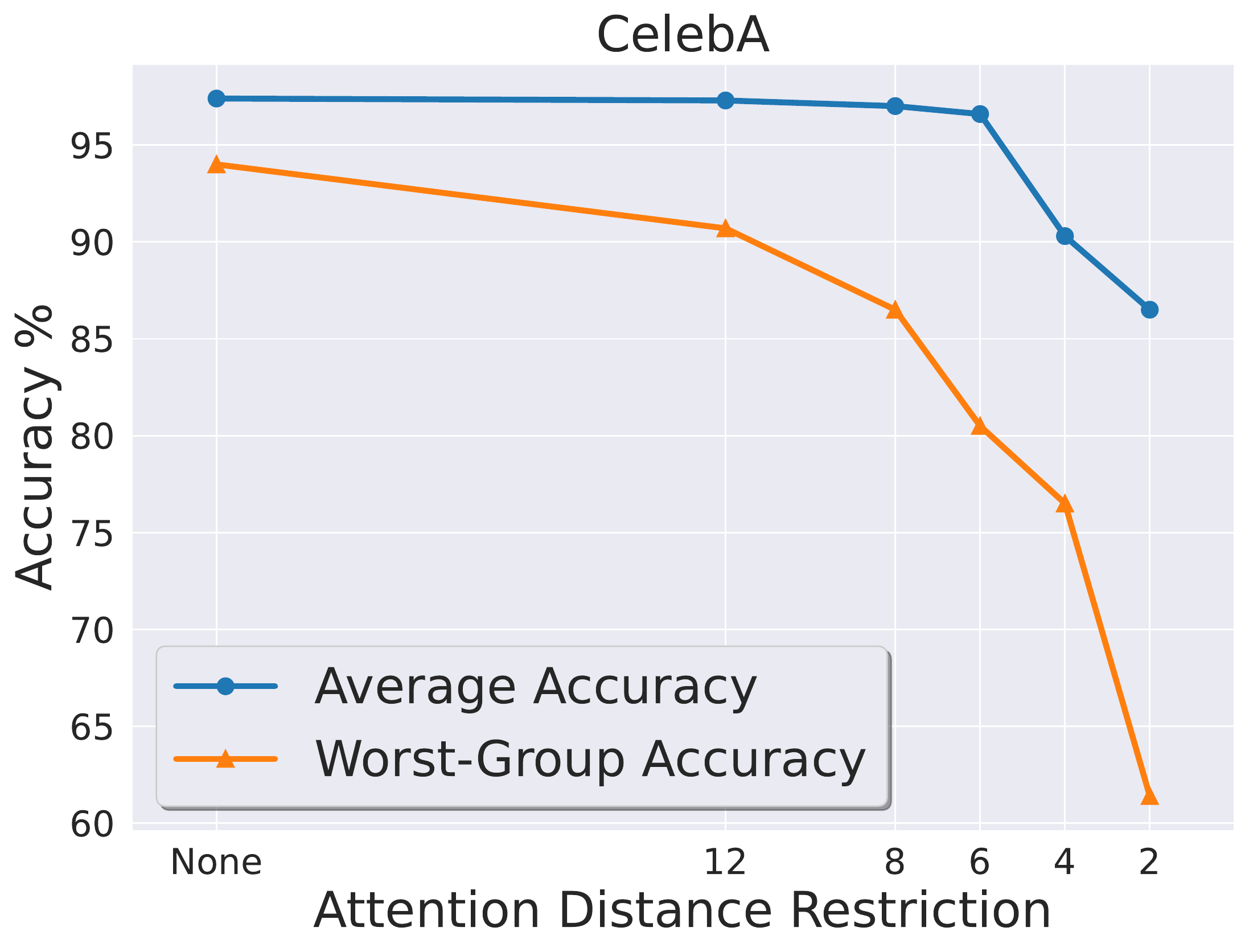}
  \label{fig:masked_attention_celeba}
\end{subfigure}
\caption{\textbf{Masked Attention.} We study the role of global attention in ViT models in providing improved robustness to spurious correlations. We observe that constraining the attention to be local results in degradation of model performance on spuriously correlated datasets such as Waterbirds (\textbf{left}) and CelebA (\textbf{right}).}
\label{fig:masked_attention}
\end{figure}

\vspace{0.4cm}
\noindent \textbf{Masked attention} The attention matrix in ViT models encapsulates crucial information about the interaction between different image patches resulting in access to more global information. Inspired by~\cite{bhojanapalli2021understanding}, we use a spatial mask to study the effect of restricting image patches to attend only those lying within a certain distance. However, the \texttt{class token} is allowed to interact and attend to all other image patches. Note, while fine-tuning we do not use any spatial mask and allow the model to leverage information from the complete attention matrix. Masking is done only during inference time. Figure~\ref{fig:masked_attention} depicts the results of our study on ViT-B/16 when fine-tuned on Waterbirds (\textbf{left}) and CelebA  (\textbf{right}). For both datasets, we see a monotonic decrease in worst-group test accuracy and Consistency Measure, as we increase the restriction on allowable attention distance. In the extreme case, when the constrained attention distance equals 2, the model completely fails to correctly classify the test images in the smallest group indicating high reliance on spurious features while making the prediction. In other words, limiting the attention to be local results in degradation of model robustness to spurious correlations. Thus, we conclude that global attention in ViT models indeed plays a crucial role in providing additional robustness to spurious correlations. 

\subsection{Investigating model performance under data imbalance}

Recall that model robustness to spurious correlations is correlated with its ability to generalize from the training examples where spurious correlations do not hold. We hypothesize that this generalization ability varies depending on the inherent data imbalance. In this section, we investigate the effect of data imbalance on the model's performance. In the extreme case, the model only observes 5 samples from the underrepresented group. 

\vspace{0.3cm}
\noindent \textbf{Setup} Considering the problem of \texttt{waterbird} vs \texttt{landbird} classification, these examples correspond to those in the groups: \texttt{waterbird} on \texttt{land} background and \texttt{landbird} on \texttt{water} background. We refer to these examples that do not include spurious associations with label as minority samples. For this study, we remove varying fraction of minority samples from the smallest group( \texttt{waterbird} on \texttt{land} background ), while fine-tuning. We measure the effect based on the worst-group test accuracy and model consistency defined in Section~\ref{sec:waterbirds}. 

 \begin{figure}[t]
\centering
\begin{subfigure}{0.5\textwidth}
  \centering
  \includegraphics[width=0.9\textwidth]{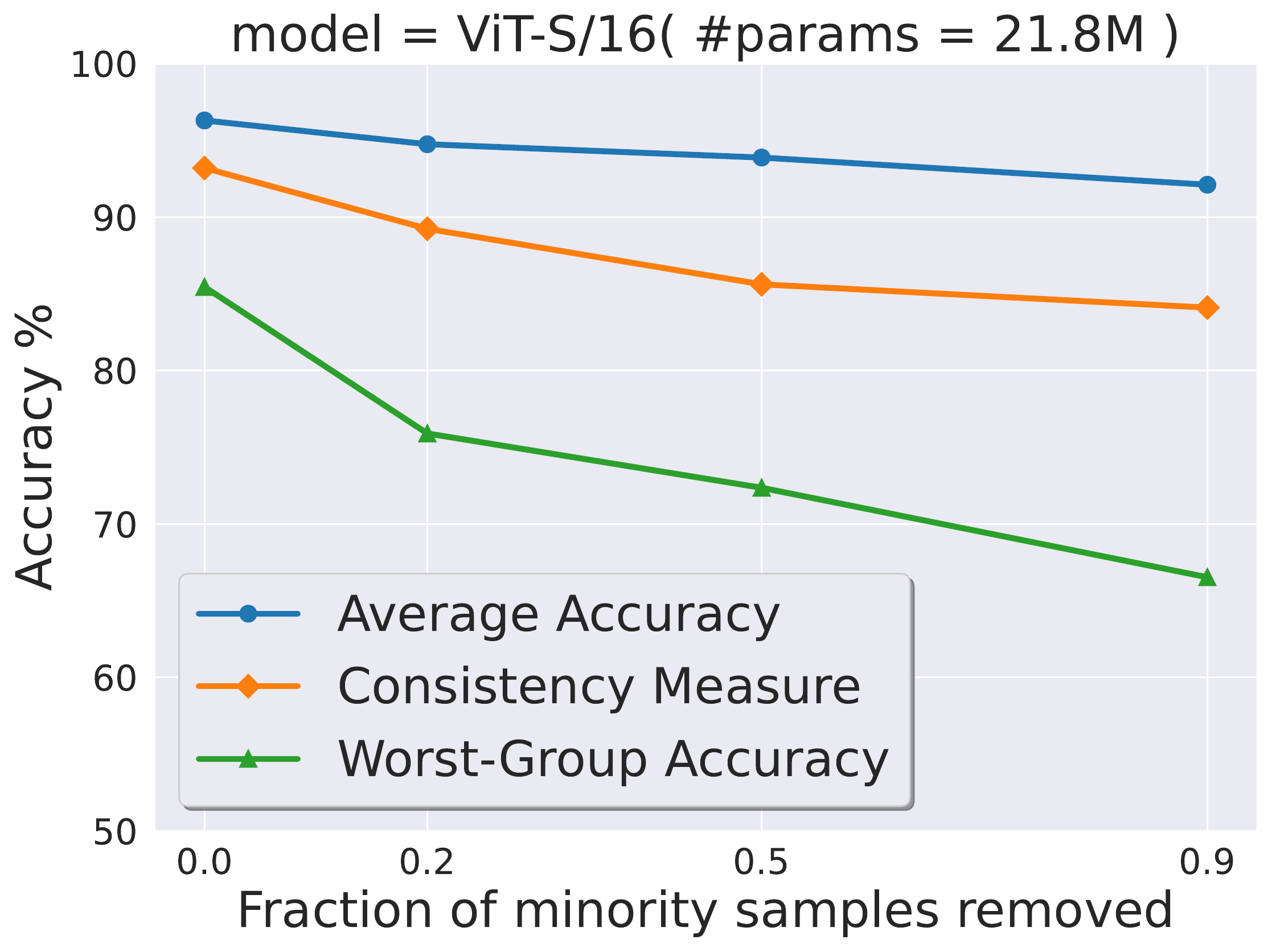}
  \label{fig:vit_data_imbalance}
\end{subfigure}%
\begin{subfigure}{0.5\textwidth}
 
  \includegraphics[width=0.9\textwidth]{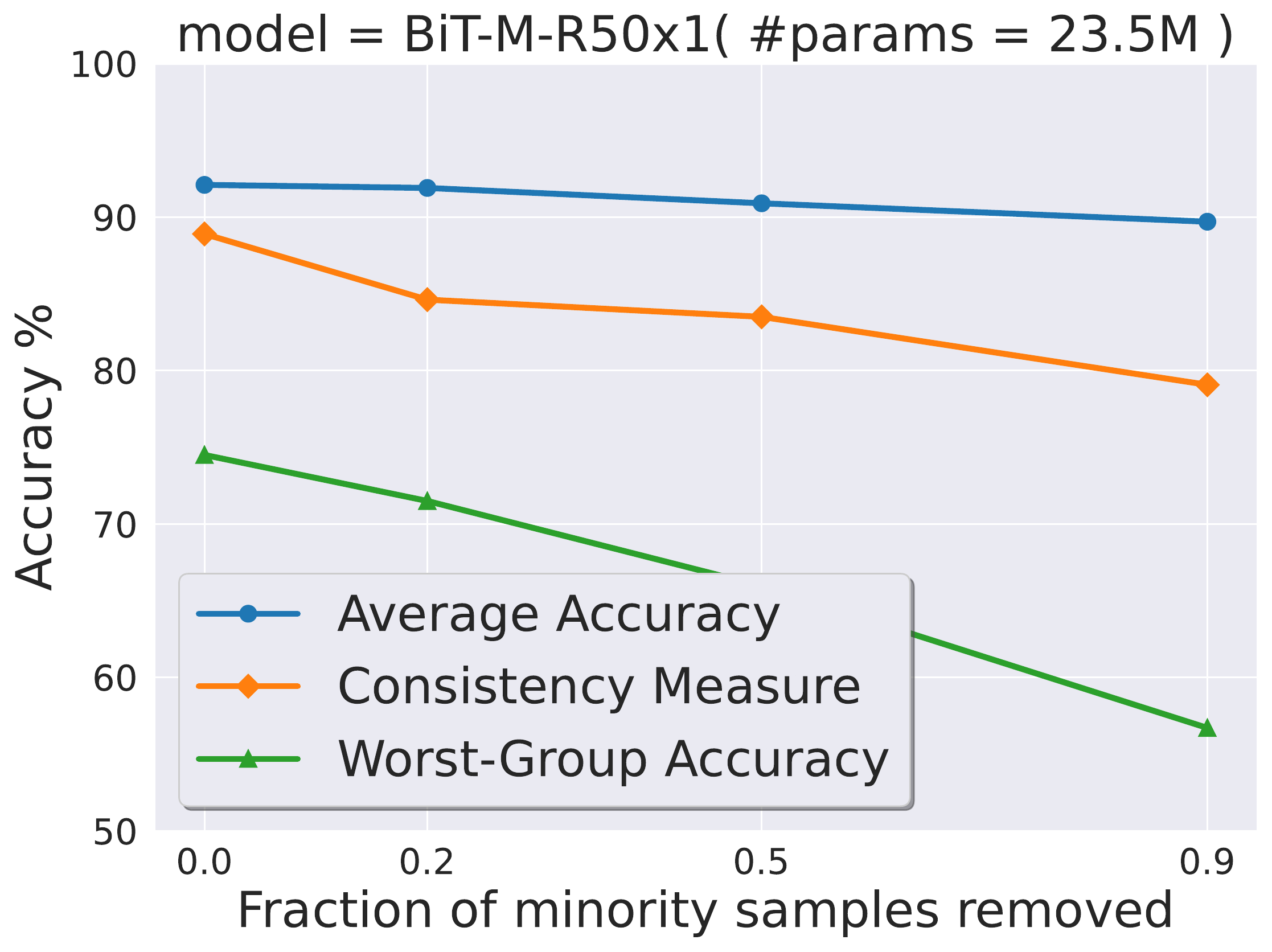}
  \label{fig:bit_data_imbalance}
\end{subfigure}
\caption{\textbf{Data Imbalance.} We investigate the effect of data imbalance on different model architectures. Our findings reveal that both ViT and BiT models suffers from spurious correlations when minority samples are scarce in fine-tuning dataset.}
\label{fig:data_imbalance}
\end{figure}

\vspace{0.3cm}
\noindent \textbf{Takeaways} In Figure~\ref{fig:data_imbalance}, we report results for ViT-S/16 and BiT-M-R50x1 model when finetuned on Waterbirds dataset~\cite{sagawa2019distributionally}. We find that as more minority samples are removed, there is a graceful degradation in the generalization capability of both ViT and BiT models. However, the decline is more prominent in BiTs with the model performance reaching near-random when we remove 90\% of minority samples. From this experiment, we conclude that additional robustness of ViT models to spurious associations stems from their better generalization capability from minority samples. However, they still suffer from spurious correlations when minority examples are scarce. 

\subsection{Does longer fine-tuning in ViT improve robustness to spurious correlations?}
Recent studies in the domain of natural language processing~\cite{lifuempirical,zhang2020revisiting} have shown that the performance of BERT~\cite{devlin2018bert} models on smaller datasets can be significantly improved through longer fine-tuning. 
In this section, we investigate if longer fine-tuning also plays a positive role in the performance of ViT models in spuriously correlated environments. 

\begin{figure}[t]
\centering
\begin{subfigure}{0.5\textwidth}
  \centering
  \includegraphics[width=0.9\textwidth]{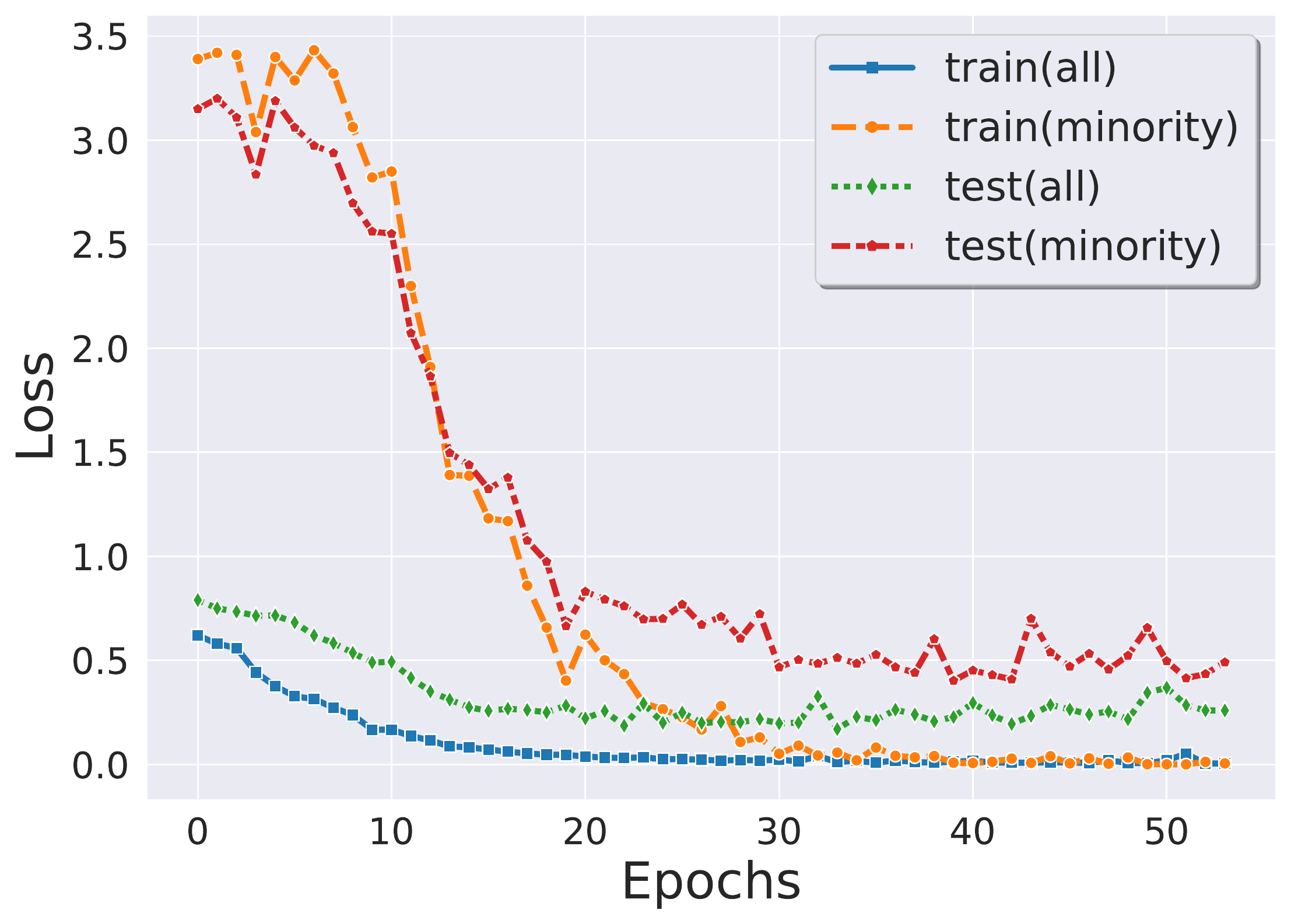}
  \label{fig:loss_plot}
\end{subfigure}%
\begin{subfigure}{0.5\textwidth}
  \includegraphics[width=0.9\textwidth]{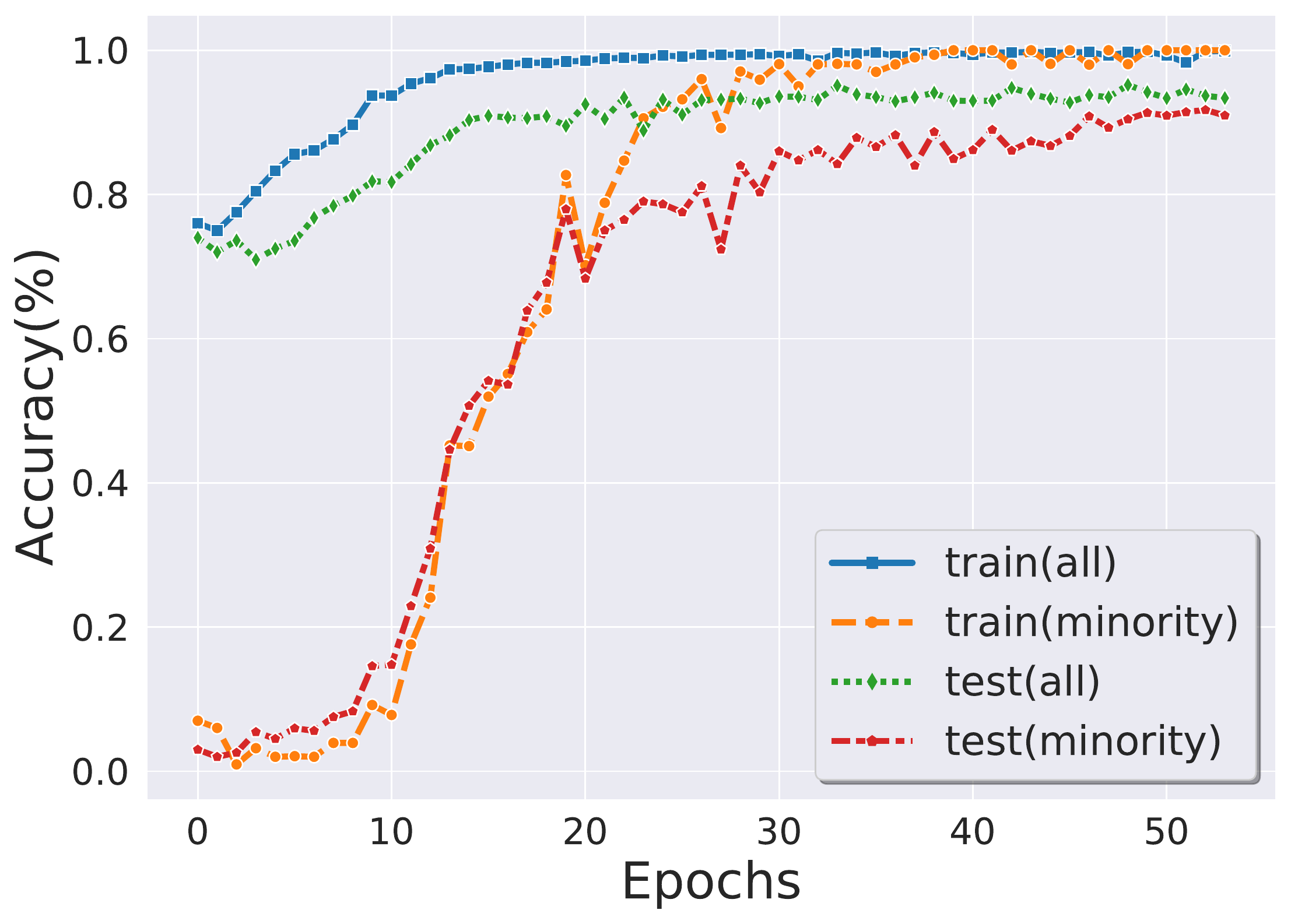}
  \label{fig:acc_plot}
\end{subfigure}
\caption{\textbf{Longer Fine-tuning.} We study the effect of longer fine-tuning on performance of ViT models. We report loss and accuracy for ViT-S/16 model finetuned on Waterbirds~\cite{sagawa2019distributionally} at each epoch of fine-tuning. Investigating further we observe that although fine-tuning for more epochs provide no additional gain in average test accuracy, but it improves model performance on minority samples.}
\label{fig:finetune_exp}
\end{figure}

\vspace{0.3cm}
\noindent \textbf{Takeaways} Figure~\ref{fig:finetune_exp} reports the loss (\textbf{left}) and accuracy (\textbf{right}) at each epoch for ViT-S/16 model fine-tuned on Waterbirds dataset~\cite{sagawa2019distributionally}. To better understand the effect of longer fine-tuning on worst-group accuracy, we separately plot the model loss and accuracy on all examples and minority samples. From the loss curve, we observe that the training loss for minority examples decreases at a much slower rate as compared to the average loss. Specifically, the average train loss takes 20 epochs of fine-tuning to reach near-zero values, while training loss on minority group plateaus after 40 epochs. Similarly, we see that although the average test accuracy of the model stops increasing after 30 epochs, the accuracy of minority samples reaches a stationary state after 50 epochs of fine-tuning. These results reveal two key observations: (1) While longer fine-tuning does not benefit the average test accuracy, it plays a positive role in improving model performance on minority samples, and (2) ViT models do not overfit with longer fine-tuning. 

\vspace{-0.2cm}
\subsection{Spurious Out-of-Distribution Detection}
\label{sec:spurious_ood}
\vspace{-0.2cm}
Finally, we study the performance of ViT models in out-of-distribution setting. Introduced in ~\cite{ming2022spurious}, spurious out-of-distribution (OOD) data is defined as samples that do not contain the invariant features $\*z^{inv}$ essential for accurate classification, but contain the spurious features $\*z^{e}$. Hence, these samples are denoted as $\*x_{ood} = \rho(\*z^{\bar{y}},\*z^e)$ where $\bar{y}$ is an out-of-class label, such that $\bar{y} \not\in \mathcal{Y}$. In the problem of \texttt{waterbird} vs \texttt{landbird} classification, an image of a person standing in forest would be an example of spurious OOD, since it contains different semantic class \texttt{person} $\not\in \{\texttt{waterbird}, \texttt{landbird}\}$, yet has the environmental features of land background. A non-robust model relying on the background feature may classify such OOD data as an in-distribution class with high confidence. 
Hence, we aim to understand if self-attention based ViT models can mitigate this problem and if so, to what extent.

\vspace{0.3cm}
\noindent\textbf{Setup}  To investigate the performance of different models against spurious OOD examples, we use the setup introduced in ~\cite{ming2022spurious}. Specifically, for Waterbirds~\cite{sagawa2019distributionally} we test on subset of images of land and water sampled from the Places dataset~\cite{zhou2017places}. Considering, CelebA~\cite{liu2015deep} as in-distribution, our test suite consists of images of \texttt{bald male} as spurious OOD, since they contain environmental features (\texttt{gender}) without invariant features (\texttt{hair}). For CMNIST, the in-distribution data contains digits $\mathcal{Y}$ = $\{0,1\}$ and the background colors, $\mathcal{E}$ = \{\texttt{red, green, purple, pink}\}. We use digits \{5, 6, 7, 8, 9\} with background color \texttt{red} and \texttt{green} as test OOD samples. 

\vspace{0.3cm}
\noindent\textbf{Takeaways} We report our findings in Table~\ref{tab:ood}. Clearly, ViT models achieve better OOD evaluation metrics as compared to BiTs. Specifically, ViT-B/16 achieves $+\textbf{32}\%$ higher AUROC than BiT-M-R50x3, considering Waterbirds~\cite{sagawa2019distributionally} as in-distribution.
 \newcolumntype{?}{!{\vrule width 1pt}}
\renewcommand{\arraystretch}{1.25}
\setlength{\tabcolsep}{4pt}
\begin{table}[h!]
\centering
 \resizebox{0.8\textwidth}{!}{
 \begin{tabular}{l|c|c?c|c?c|c}
 \hlineB{2.5}
 \multirow{2}{0.75cm}{\centering Model} & \multicolumn{2}{c?}{Waterbirds~\cite{sagawa2019distributionally}} & \multicolumn{2}{c?}{CelebA~\cite{liu2015deep}} &
 \multicolumn{2}{c}{CMNIST} \\
 \cline{2-7}
 & FPR95$\pmb{\downarrow}$ & AUROC$\pmb{\uparrow}$ & FPR95$\pmb{\downarrow}$ & AUROC$\pmb{\uparrow}$ & FPR95$\pmb{\downarrow}$ & AUROC$\pmb{\uparrow}$ \\
  \hline
ViT-B/16 & \textbf{56.8} & \textbf{91.0} & \textbf{60.5} & \textbf{88.4} & \textbf{7.4} & \textbf{98.8} \\
 \hline
 ViT-S/16 & 62.2 & 87.0 & 61.3 & 86.7 & 8.7 & 97.7 \\ 
 \hline
 ViT-Ti/16 & 79.5 & 71.6 & 94.3 & 72.7 & 16.4 & 96.7\\ 
 \hline
 BiT-M-R50x3 & 96.0 & 59.0 & 63.8 & 85.3 & 45.9 & 84.1 \\ 
 \hline
BiT-M-R101x1 & 95.5 & 59.5 & 70.3 & 85.6 & 44.5 & 81.4 \\ 
 \hline
 BiT-M-R50x1 & 95.1 & 63.4 & 69.7 & 85.7 & 30.0 & 88.4 \\ 
 \hlineB{2.5}
 \end{tabular}}
 \vspace{0.25cm}
\caption{\textbf{Spurious OOD evaluation.} OOD detection performance of ViT and BiT models when finetuned on Waterbirds~\cite{sagawa2019distributionally}, CelebA~\cite{liu2015deep} \& CMNIST. We use energy score~\cite{liu2020energy} for calculating AUROC and FPR95. We observe that ViT models are more robust to spurious OOD examples as compared to BiTs.}
\label{tab:ood}
\end{table}

\section{Related Works}
\label{sec:related_works}
\noindent\textbf{Pre-training and robustness} 
Recently, there has been an increasing amount of interest in studying the effect of pre-training~\cite{Kolesnikov2020BigLearning, devlin2018bert, radford2021learning, liu2019roberta}. 
Specifically, when the target dataset is small, generalization can be significantly improved through pre-training and then finetuning~\cite{zeiler2014visualizing}. Findings of Hendrycks \textit{et   al.}~\cite{hendrycks2019using} reveal that pre-training provides significant improvement to model robustness against label corruption, class imbalance, adversarial examples, out-of-distribution detection, and confidence calibration. In this work, we focus distinctly on robustness to \emph{spurious correlation}, and how it can be improved through large-scale pretraining. 

\vspace{0.2cm}
\noindent\textbf{Vision transformer} Since the introduction of transformers by Vaswani \textit{et al.}~\cite{vaswani2017attention} in 2017, there has been a deluge of studies adopting the attention-based transformer architecture for solving various problems in natural language processing~\cite{radford2018improving,radford2019language,yang2019xlnet,dai2019transformer}. In the domain of computer vision,  Dosovitskiy \textit{et al.}~\cite{dosovitskiy2020image} first  introduced the concept of Vision Transformers (ViT) by adapting the transformer architecture in \cite{vaswani2017attention} for image classification tasks. Subsequent studies~\cite{dosovitskiy2020image,steiner2021train} have shown that when pre-trained on sufficiently large datasets, ViT achieves superior performance on downstream tasks, and outperforms state-of-art CNNs such as residual networks (ResNets)~\cite{He2016DeepRecognition} of comparable sizes. 
Since coming to the limelight, multiple variants of ViT models have been proposed. Touvron \textit{et al.}~\cite{touvron2021training} showed that it is possible to achieve comparable performance in small pre-training data regimes using extensive data augmentation and novel distillation strategy. Further improvements on ViT include enhancement in tokenization module~\cite{yuan2021tokens}, efficient parameterization for scalability~\cite{touvron2021going, xue2021go, zhai2021scaling} and  building multi-resolution feature maps on transformers~\cite{liu2021swin, wang2021pyramid}. In this paper, we provide a first systematic study on the robustness of vision transformers when learned on datasets containing spurious correlations. 

\vspace{0.3cm}
\noindent\textbf{Robustness of transformers} Naseer \textit{et al.}~\cite{naseer2021intriguing} provides a comprehensive understanding of the working principle of ViT architecture through extensive experimentation. Some notable findings in \cite{naseer2021intriguing} reveal that transformers are highly robust to severe occlusions, perturbations, and distributional shifts.  Recently, performance of ViT models in the wild has been extensively studied~\cite{bhojanapalli2021understanding,zhang2021delving, paul2021vision, bai2021transformers} using a set of robustness generalization benchmarks, e.g., ImageNet-C~\cite{hendrycks2018benchmarking}, Stylized-ImageNet~\cite{geirhos2018imagenet}, ImageNet-A~\cite{hendrycks2021natural}, etc. Different from prior works, we focus on robustness performance on challenging datasets, which are designed to expose spurious correlations learned by the model. Our analysis reveals that pre-training improves robustness by better generalizing on examples from under-represented groups. Our findings are also complementary to robustness studies~\cite{lifuempirical,he2019unlearn,mccoy2019right} in the domain of natural language processing, which reported that transformer-based BERT~\cite{devlin2018bert} models improve robustness to spurious correlations.

\section{Conclusion}
\label{sec:conclusion}
In this paper, we investigate the robustness of ViT models when learned on datasets containing spurious associations between target label and environmental features. Our findings can be summarized as: 1) ViTs are more robust to spurious correlations than CNNs under large-scale pre-training data regime. However, when the pre-training dataset is relatively small, transformer models perform much worse as compared to CNNs of comparable size; 2) We find that global attention in ViT architecture plays a crucial role in providing improved robustness. Further, restricting the attention to be local results in degradation of model performance; 3) Improved robustness of ViT models can be attributed to better generalization capability from the counterexamples where spurious correlations do not hold. However, when such samples become scarce ViT models tend to overfit to spurious associations. We hope that our work will inspire future research on understanding the robustness of ViT models.
\clearpage
\bibliographystyle{plain}
\bibliography{egbib}

\newpage
\appendix
\begin{center}
    \Large{\textbf{Supplementary Material}}
\end{center}
\section{Implementation Details}
\begin{enumerate}
    \item \textbf{Transformers.} For both ViT and DeiT models, we obtain the pre-trained checkpoints from the \texttt{timm} library\footnote{\url{https://github.com/rwightman/pytorch-image-models/tree/master/timm}}. For downstream fine-tuning on Waterbirds and CelebA dataset, we scale
up the resolution to 384 × 384 by adopting 2D interpolation of the pre-trained position embeddings proposed in \cite{dosovitskiy2020image}. Note, for CMNIST we keep the resolution as $224 \times 224$ during fine-tuning. We fine-tune models using SGD with a momentum of 0.9 with an initial learning rate of 3e-2. As described in \cite{steiner2021train}, we
use a fixed batch size of 512, gradient clipping at global norm 1 and a cosine decay learning rate schedule with a linear warmup. We fine-tune \texttt{tiny} \& \texttt{small} versions of models (\emph{i.e.,} ViT-Ti/16 and ViT-S/16) for 1000 steps, whereas \texttt{base} version (\emph{i.e.,} ViT-B/16) is fine-tuned for 2000 steps.
\vspace{0.3cm}
    \item \textbf{BiT.} We obtain the pretrained checkpoints from the official repository\footnote{\url{https://github.com/google-research/big_transfer}}. For downstream fine-tuning, we use SGD with an initial learning rate of 0.003, momentum 0.9, and batch size 512. We fine-tune models with various capacity for 500 steps, including BiT-M-R50x1, BiT-M-R50x3, and BiT-M-R101x1.
\end{enumerate}

\section{Extension: How does the size of pre-training dataset affect robustness to spurious correlations?}
In this section, to further validate our findings on the importance of large-scale pre-training dataset, we show results on CelebA~\cite{liu2015deep} dataset. We report our findings in Table~\ref{tab:pretraining_vit_celeba}. We also observe a similar trend for this setup that larger model capacity and more pre-training data yields significant improvement in worst-group accuracy for ViT models. Further, when pre-trained on a relatively smaller dataset such as ImageNet-1k, the performance of transformer-based DeiT models are poor as compared to the corresponding CNN counterpart.

Also, compared to BiT models, \emph{the robustness of ViT models benefits more with a large pre-training dataset}. For example, compared to ImageNet-1k, fine-tuning ViT-B/16 pre-trained on ImageNet-21k improves the worst-group accuracy by \textbf{6\%}. On the other hand, for BiT models, fine-tuning with a larger pre-trained dataset yields marginal improvement. Specifically, BiT-M-R50x3 only improves the worst-group accuracy by 1.5\% with ImageNet-21k. 
\newcolumntype{?}{!{\vrule width 1pt}}
\renewcommand{\arraystretch}{1.25}
\setlength{\tabcolsep}{6pt}
\begin{table}[t]
\centering
\resizebox{0.7\textwidth}{!}{
 \begin{tabular}{cl|c?c}
 \hlineB{2.5}
 & \multirow{2}{0.75cm}{\centering Model} & \multicolumn{2}{c}{Test Accuracy} \\
 \cline{3-4}
&  & Average Acc. & Worst-Group Acc.  \\
  \hline
\multirow{3}{2cm}{ImageNet-21k} & ViT-B/16 & \textbf{97.4} & \textbf{94.0} \\

& ViT-S/16 & 97.0 & 91.5 \\ 

& ViT-Ti/16 & 96.5 & 84.6\\ 
 \hlineB{2}
\multirow{3}{2cm}{ImageNet-1k} & DeiT-B/16 & 96.4 & 88.0 \\ 

& DeiT-S/16 & 96.1 & 87.1\\ 

& DeiT-Ti/16 & 94.9 & 85.7 \\ 
 \hlineB{2}
  \multirow{3}{2cm}{ImageNet-21k} & BiT-M-R50x3 & 97.3 & 89.8 \\
& BiT-M-R101x1 & 97.2 & 89.8 \\ 
& BiT-M-R50x1 & 96.8 & 87.7 \\ 
 \hlineB{2}
 \multirow{3}{2cm}{ImageNet-1k}& BiT-S-R50x3 & 96.4 & 88.3\\ 
& BiT-S-R101x1 & 96.5 & 90.2 \\ 
& BiT-S-R50x1 & 96.3 & 90.9 \\ 
 \hlineB{2.5}\\
 \end{tabular}}
\caption{Investigating the effect of large scale pre-training on model robustness to spurious correlations when finetuned on CelebA~\cite{liu2015deep}. }
\label{tab:pretraining_vit_celeba}
\end{table}

\section{Extension : Color Spurious Correlation}
\label{sec:cmnist}
  To further validate our findings beyond natural background and gender as spurious (\emph{i.e.} environmental) features, we provide additional experimental results with the ColorMNIST dataset, where the digits are superimposed on coloured backgrounds. Specifically, it contains spurious correlation between the target label and the background color. Similar to the setup in~\cite{ming2022spurious}, we fix the classes $\mathcal{Y}$ = $\{0,1\}$ and the background colors, $\mathcal{E}$ = \{\texttt{red, green, purple, pink}\}. For this study, label $y=0$ is spuriously correlated with background color $\{\texttt{red}, \texttt{purple}\}$, and similarly, label $y=1$ has spurious associations with background color $\{\texttt{green}, \texttt{pink}\}$. Formally, we have $\dP(e = \texttt{red} | y = 0) = \dP(e = \texttt{purple} | y = 0) = \dP(e = \texttt{green} | y = 1) = \dP(e = \texttt{pink} | y = 1) = 0.45$ and $\dP(e = \texttt{green} | y = 0) = \dP(e = \texttt{pink} | y = 0) = \dP(e = \texttt{red} | y = 1) = \dP(e = \texttt{purple} | y = 1) = 0.05$.  Note that, while fine-tuning the models, we fix the foreground color of digits as $\texttt{white}$.
  
\begin{figure}[h!]
    \centering
    \includegraphics[scale = 0.5]{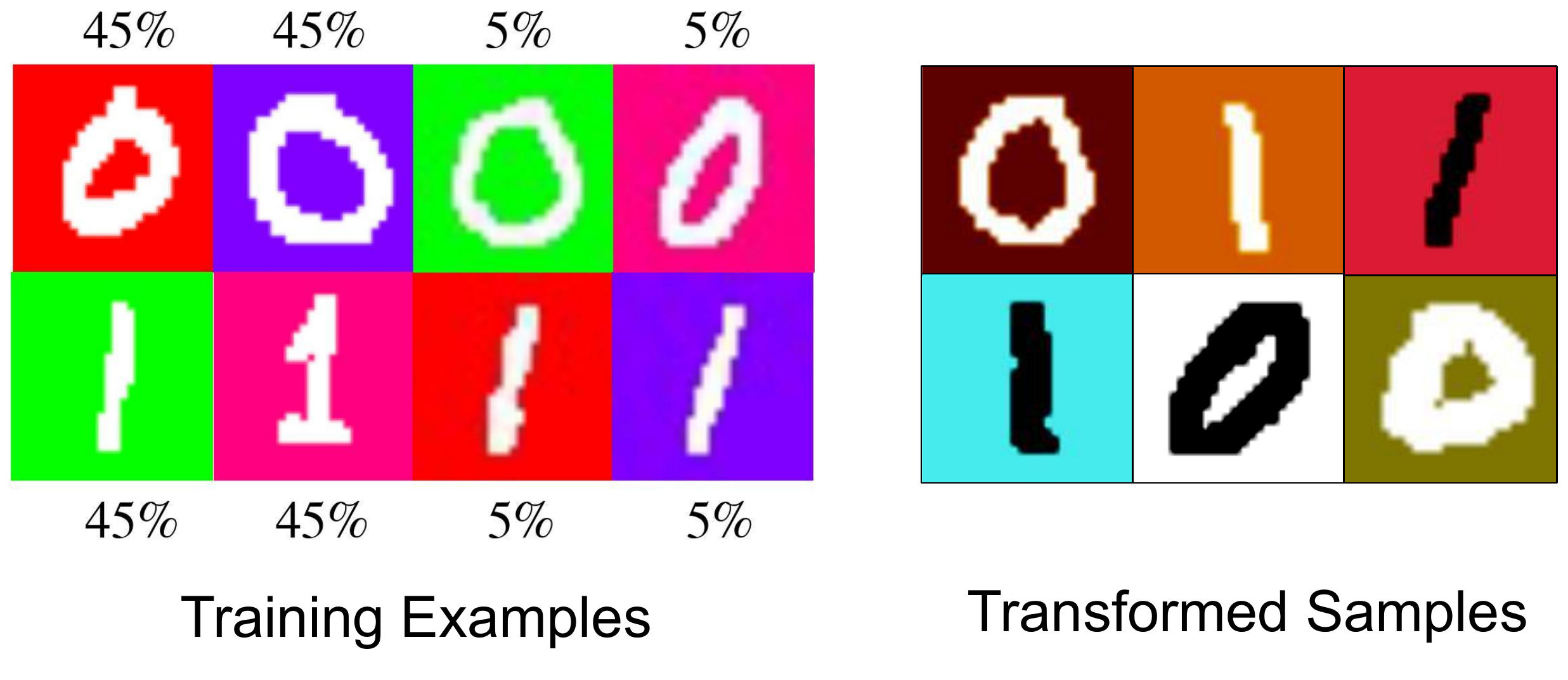}
    \caption{\textbf{CMNIST.} Distribution of training samples in CMNIST dataset(\textbf{left}) and few representative examples after  transformation(\small\textbf{right}) as defined in Section~\ref{sec:cmnist}.}
    \label{fig:cmnist_setup}
\end{figure}

  \vspace{0.3cm}
  \noindent \textbf{Results  and  insights  on  robustness  performance} We compare model predictions on samples with same class label but different background \& foreground colors. Given a data point ($\*x_i, y_i$), we modify the background and foreground color of $\*x_i$ randomly to generate a new test image $\bar{\*x}_i$ with the constraint of having the same semantic label. During evaluation, the background color is chosen uniform-randomly from the set of colors: \texttt{\{\#ecf02b, \#f06007, \#0ff5f1, \#573115, \#857d0f, \#015c24, \#ab0067, \#fbb7fa, \#d1ed95, \#0026ff\}} and the foreground color is selected randomly from the set $\{\texttt{black},\texttt{white}\}$. For evaluation purpose, we form a dataset consisting of $2100$ samples and the results reported are averaged over 50 random runs. Figure~\ref{fig:cmnist_setup} depicts the distribution of training samples in CMNIST dataset (\textbf{left}) and few representative examples after transformation (\textbf{right}).
 
\begin{figure}[h!]
\centering
\begin{subfigure}{0.55\textwidth}
  \centering
  \includegraphics[width=0.85\textwidth]{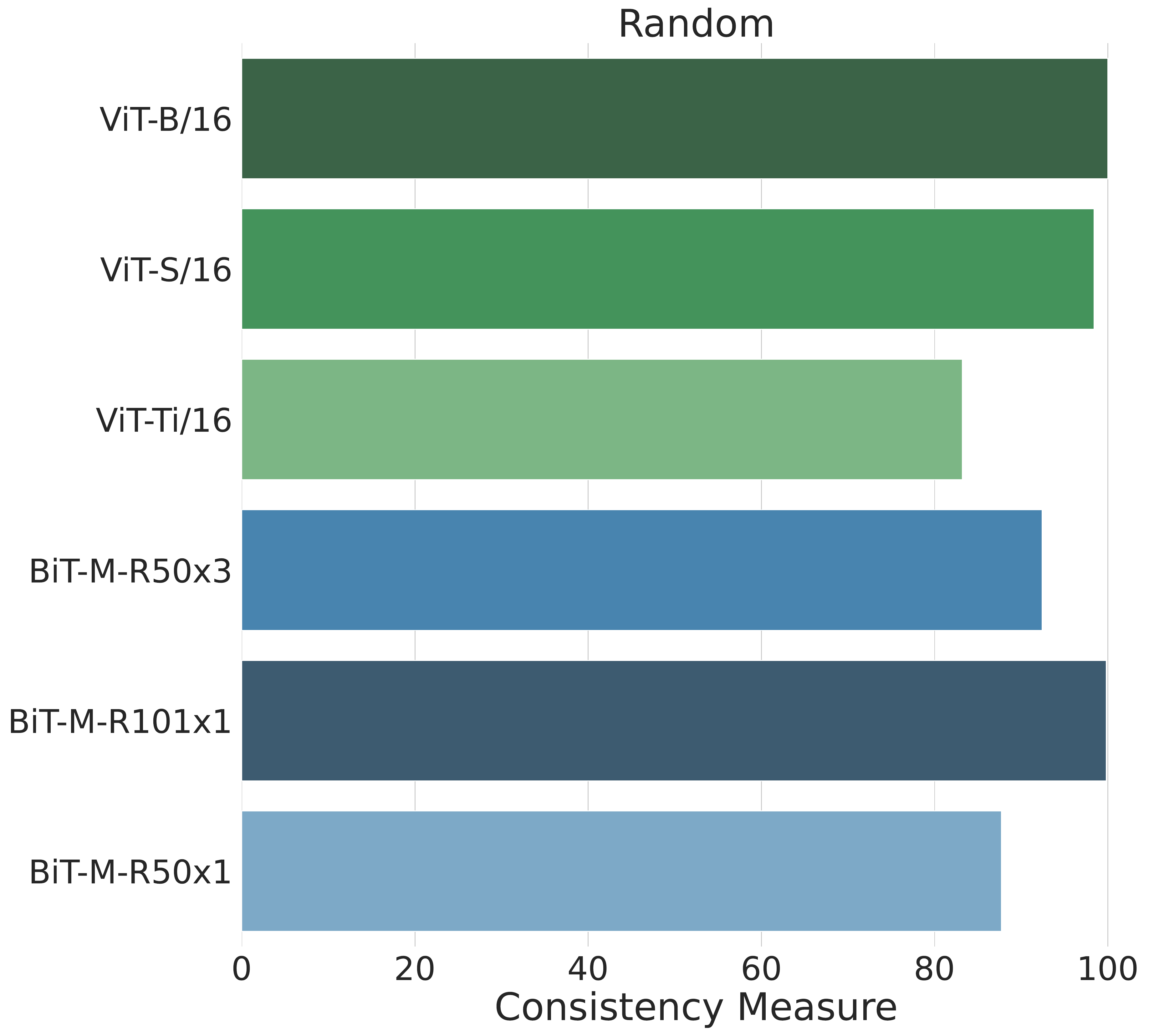}
\end{subfigure}%
\begin{subfigure}{0.55\textwidth}
  \includegraphics[width=0.85\textwidth]{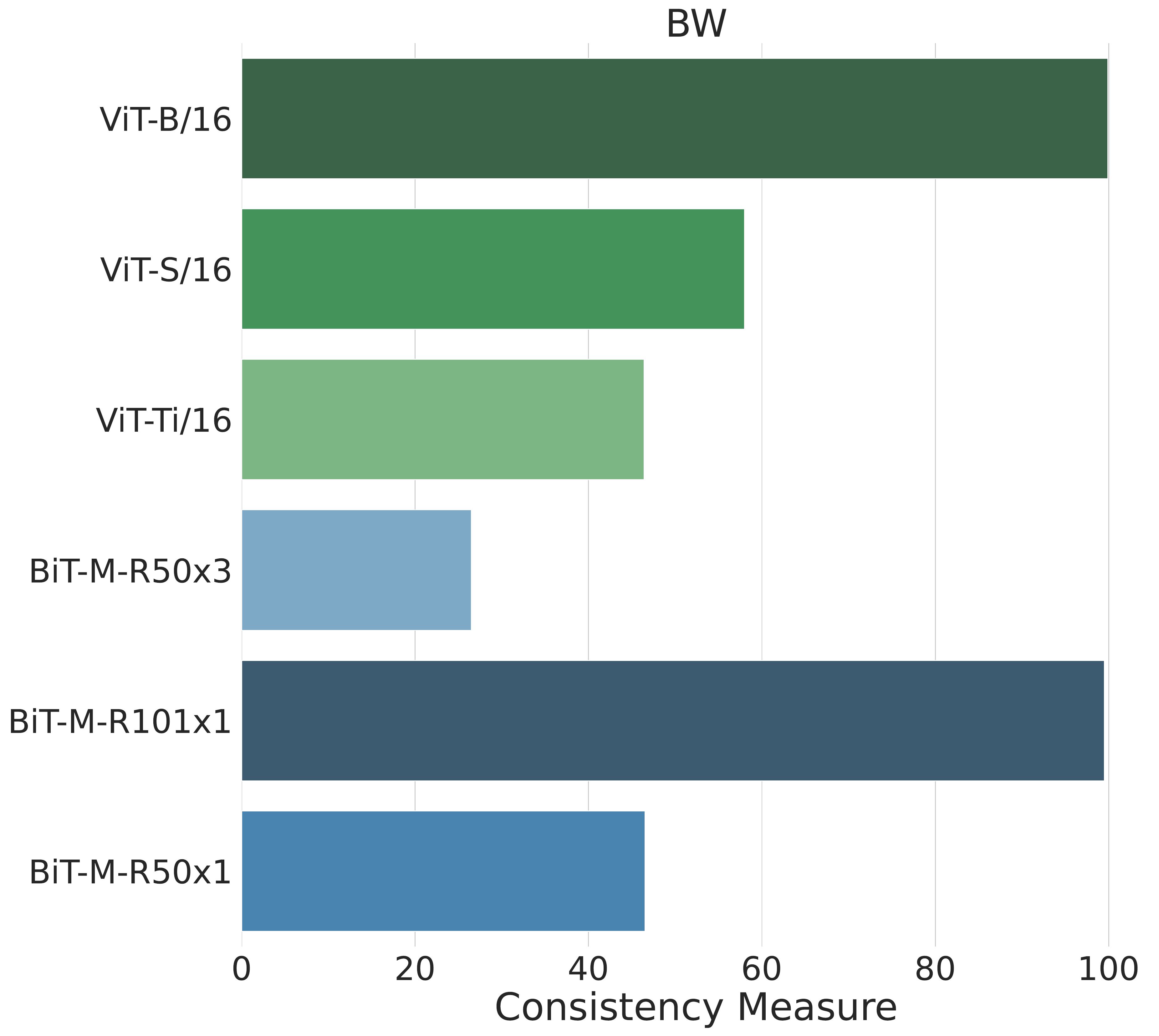}
\end{subfigure}

\caption{\textbf{Consistency Measure.} Evaluation results
quantifying consistency for models of different architectures and varying capacity. We indicate the setup when the foreground color is set as $\texttt{black}$ and the background as $\texttt{white}$ using \textbf{BW}(\textbf{right}). \textbf{Random} represents setting both the foreground and background color randomly(\textbf{left}).} 
\label{fig:cmnist_robustness_plot}
\end{figure}

  \vspace{0.3cm}
  We report our findings in Figure~\ref{fig:cmnist_robustness_plot}. Our operating hypothesis is that a robust model should  predict same class label $\hat{f}(\*x_i)$ and ${\hat{f}}(\bar{\*x}_i)$  for a given pair $(\*x_i, \bar{\*x}_i)$, as they share exactly the same target label (i.e., the invariant feature is approximately the same). We can observe from Figure~\ref{fig:cmnist_robustness_plot} that the best model ViT-B/16 obtains consistent predictions for 100\% of image pairs.   After extensive experimentation over all combinations, we find that setting the foreground color as $\texttt{black}$ and the background as $\texttt{white}$ caused the models to be most vulnerable. We see a significant decline in model consistency when the foreground color is set as $\texttt{black}$ and the background as $\texttt{white}$ (indicated as \textbf{BW}) as compared to random setup.
  
\section{Visualization}
\subsection{Attention Map}
\begin{figure}[h!]
\centering
\begin{subfigure}{0.5\textwidth}
  \centering
  \includegraphics[width=0.8\textwidth]{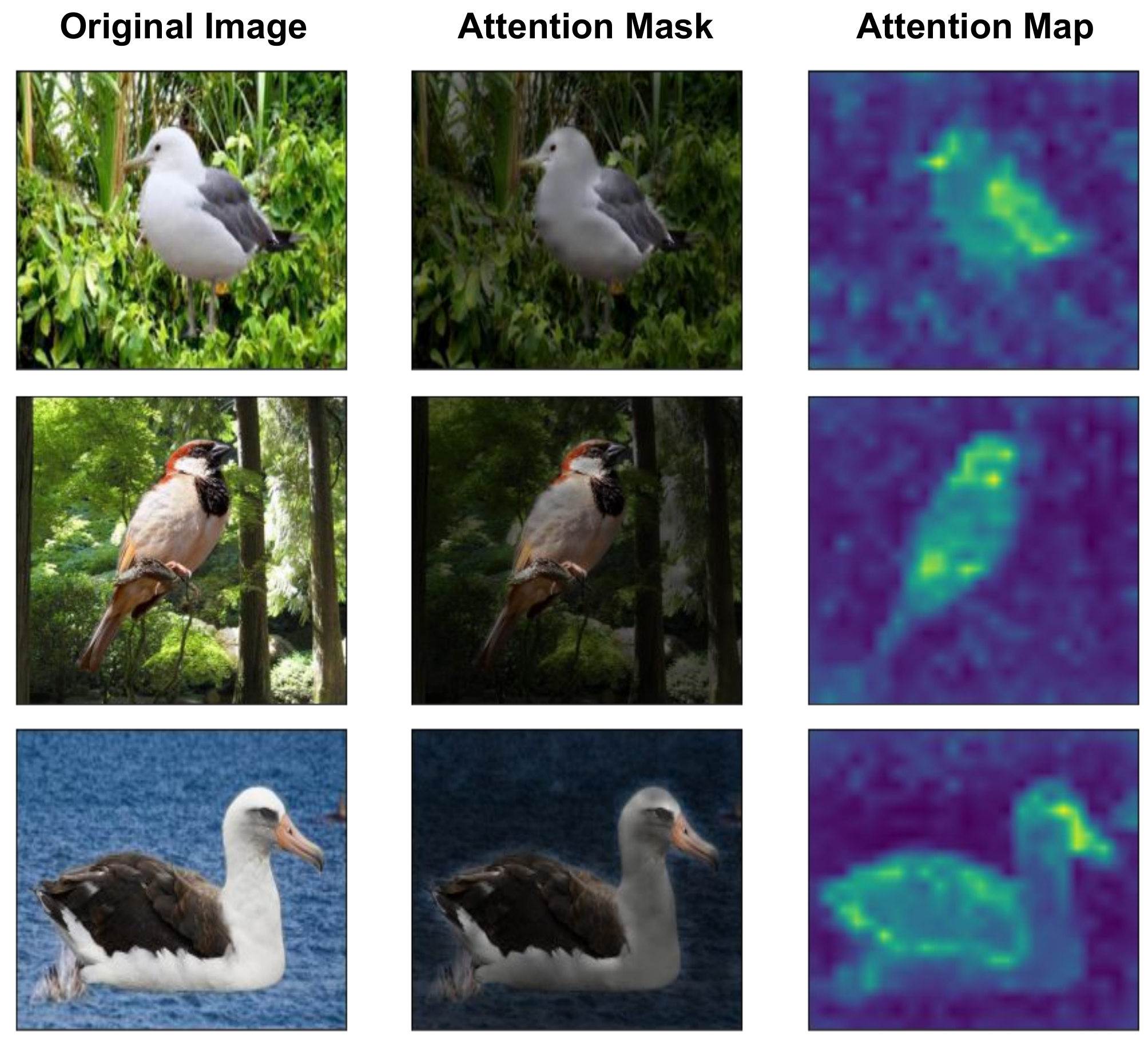}
\end{subfigure}%
\begin{subfigure}{0.5\textwidth}
  \includegraphics[width=0.8\textwidth]{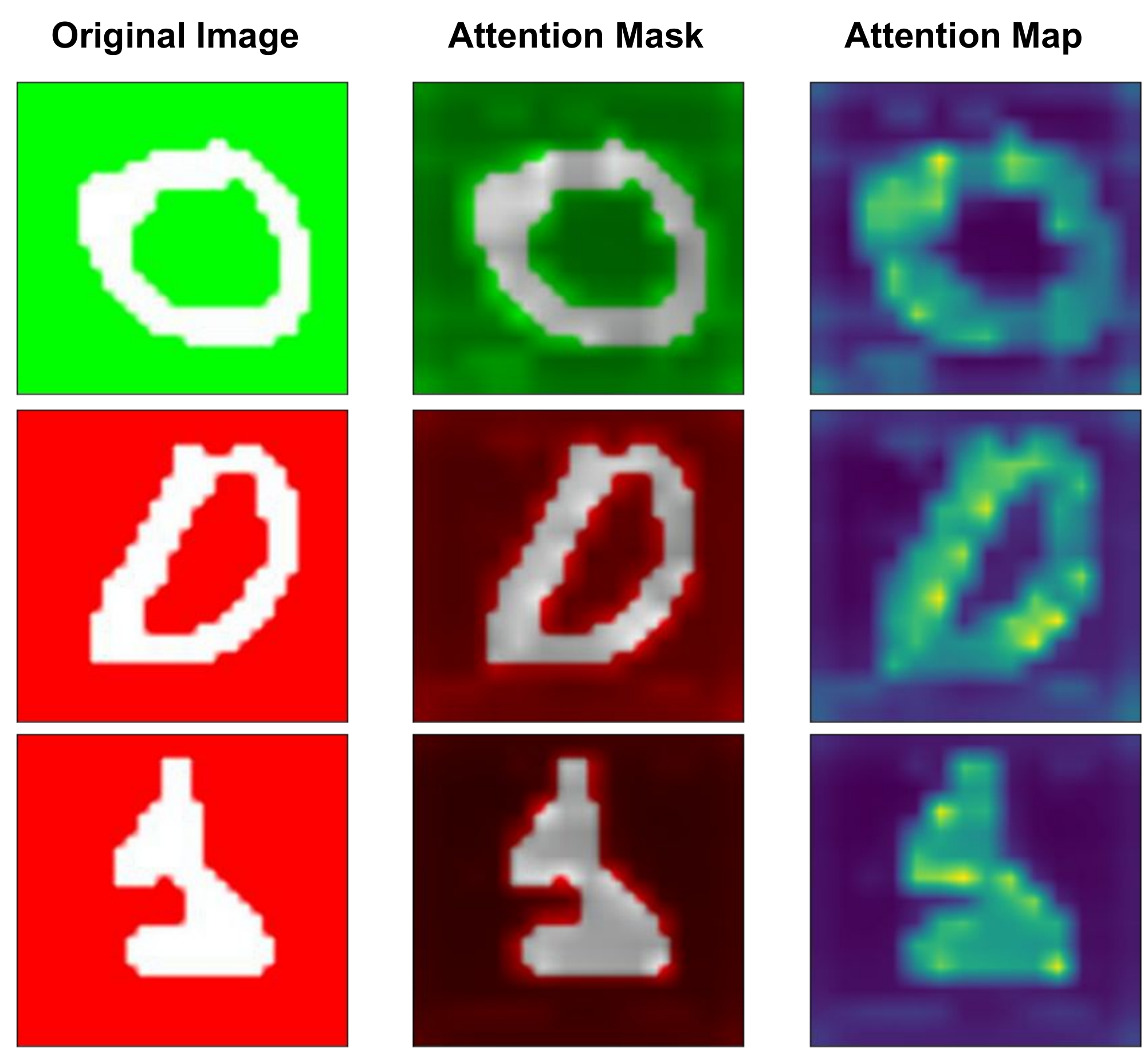}
\end{subfigure}
\caption{\textbf{Attention Map.} Visual illustration of attention map obtained from ViT-B/16 model for few representative images.} 
\label{fig:attention_mask}
\end{figure}

In Figure~\ref{fig:attention_mask}, we visualize attention maps obtained from ViT-B/16 model for some samples images from Waterbirds~\cite{sagawa2019distributionally} and CMNIST dataset. We use Attention Rollout~\cite{abnar-zuidema-2020-quantifying} to obtain the attention matrix. We can observe that the model successfully attends spatial locations representing invariant features while making predictions.

\subsection{The Attention Matrix of CMNIST}
\begin{figure}[t]
    \centering
    \includegraphics[scale = 0.5]{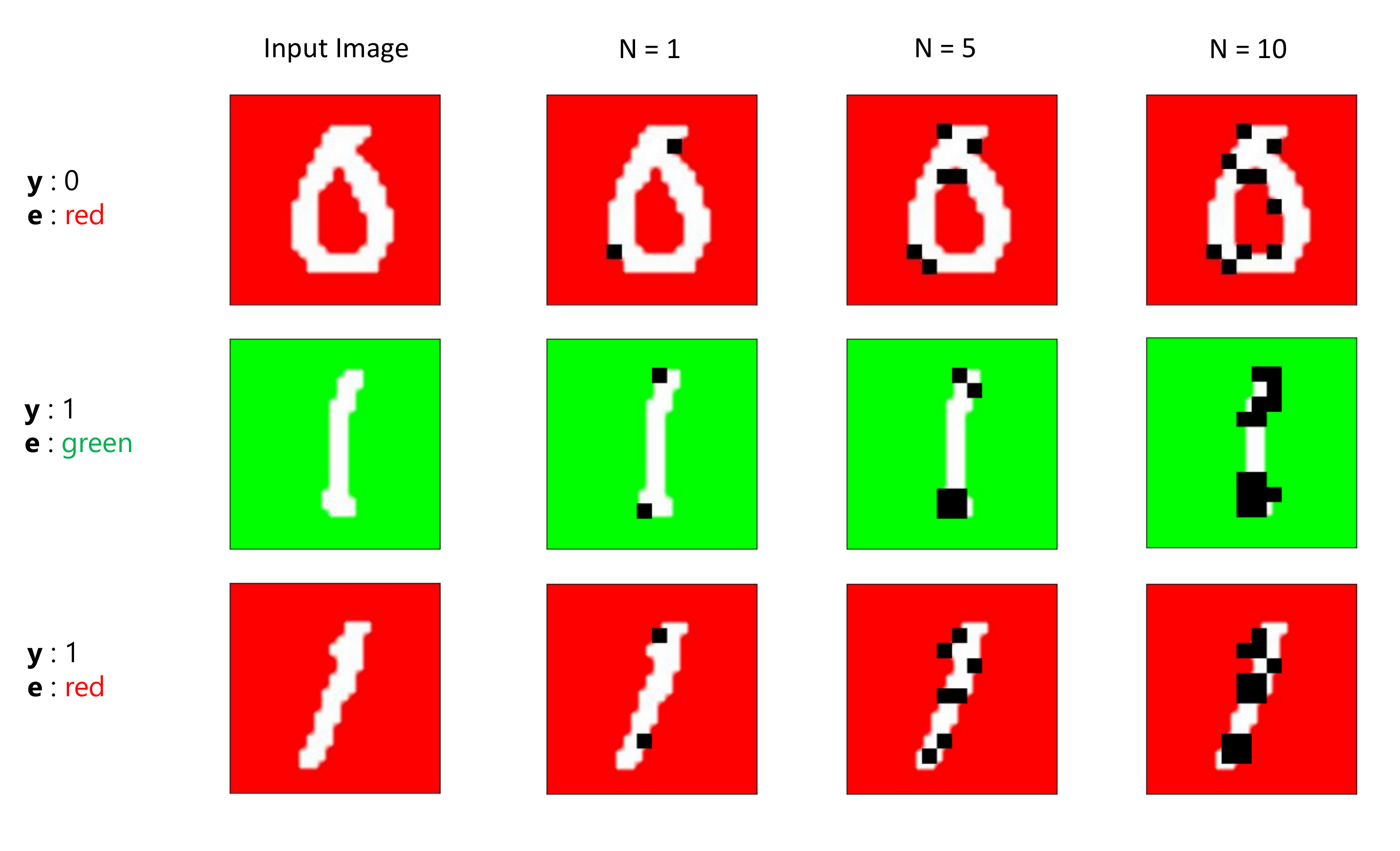}
    \caption{Visualization of the top N patches receiving the highest attention (marked in \textcolor{black}{\textbf{black}}) for ViT-B/16 fine-tuned on CMNIST. Investigating the attention matrix, we find that all image patches—irrespective of spatial location—provides maximum attention to the patches representing essential cues}
    \label{fig:cmnist_back_patch}
\end{figure}

In the main text, we provide visualizations in which each image patch, irrespective of its spatial location, provides maximum attention to the patches representing essential cues for accurately identifying the foreground object. In Figure~\ref{fig:cmnist_back_patch}, we show visualizations for ViT-B/16 fine-tuned on CMNIST dataset to further validate our findings.

\end{document}